\documentclass{article}

\usepackage[preprint]{neurips_2019}

\usepackage[utf8]{inputenc} 
\usepackage[T1]{fontenc}    
\usepackage{hyperref}       
\usepackage{url}            
\usepackage{booktabs}       
\usepackage{amsfonts}       
\usepackage{nicefrac}       
\usepackage{microtype}      

\usepackage{amsmath}
\usepackage[pdftex]{graphicx}
\usepackage[flushleft]{threeparttable}

\usepackage{algorithm}  
\usepackage{algpseudocode}  

\usepackage{graphicx}

\usepackage{multirow}

\title{Explicitly Bayesian Regularizations in Deep Learning}

\author{%
  Xinjie Lan, Kenneth E. Barner\\
  Department of Electrical and Computer Engineering\\
  University of Delaware\\
  Newark, DE 19713 \\
  \texttt{lxjbit@udel.edu} \\
}

\begin{document}

\maketitle

\begin{abstract}


Generalization is essential for deep learning.
In contrast to previous works claiming that Deep Neural Networks (DNNs) have an implicit regularization implemented by the stochastic gradient descent, we demonstrate explicitly Bayesian regularizations in a specific category of DNNs, i.e., Convolutional Neural Networks (CNNs).
First, we introduce a novel probabilistic representation for the hidden layers of CNNs and demonstrate that CNNs correspond to Bayesian networks with the serial connection.
Furthermore, we show that the hidden layers close to the input formulate prior distributions, thus CNNs have explicitly Bayesian regularizations based on the Bayesian regularization theory.
In addition, we clarify two recently observed empirical phenomena that are inconsistent with traditional theories of generalization.
Finally, we validate the proposed theory on a synthetic dataset.

\end{abstract}

\section{Introduction}

Generalization is a fundamental problem of deep learning theory.
Since traditional generalization theories cannot clarify over-parameterized Deep Neural Networks (DNNs), especially in some cases, e.g., the random label classification \cite{generalization_regularization, generalization_regularization1, regularization5},
most recent works ascribe the generalization performance of DNNs to the Stochastic Gradient Descent (SGD) optimization implementing an implicit regularization, i.e., a bias towards models with low complexity \cite{keskar2016large, dinh2017sharp, generalization_regularization2, soudry2018implicit, arora2019implicit}.

Nevertheless, the implicit regularization explanations have lots of controversies.
For example, Keskar et al. claim that the sharp minima learned by SGD lead to poor generalization \cite{keskar2016large}, but Dinh et al. demonstrate that sharp minima can generalize DNNs well \cite{dinh2017sharp}.
Sourdry et al. argue that SGD learns the minimum norm as the implicit regularization \cite{generalization_regularization2, soudry2018implicit}, but Arora et al. doubt that the implicit regularization cannot be captured by the mathematical norm \cite{arora2019implicit, generalization_regularization1}.

In the context of regularization theories, Bayesian theory indicates that an explicit regularization can be formulated as a prior distribution \cite{bayesian_regularization}.
However, it is scarce that explaining the regularization of DNNs from the Bayesian viewpoint, because the extremely complicated architecture of DNNs and the difficulty of probabilistic modeling high-dimensional activations impede establishing a comprehensive probabilistic explanation for the architecture of DNNs. 

Numerous efforts have been devoted to explain the architecture of DNNs from a probabilistic perspective \cite{deep-BM, Boltzmann_machine, DGMM, DMFA, DRMM, DNN_GP, Mattews_GP, garriga2018deep, CNN_GP2}. However, existing probabilistic explanations have several limitations. 
The Boltzmann explanations only clarify the restricted Boltzmann machine \cite{deep-BM, Boltzmann_machine}.
The mixture model explanations require too strict assumptions \cite{DGMM, DMFA, DRMM}.
The gaussian process explanations cannot clarify the hierarchical representation, an essential of deep learning \cite{Mattews_GP}.
In summary, existing explanations cannot establish a solid probabilistic foundation for investigating the regularization of DNNs from the Bayesian perspective.

\pagebreak
To establish a solid probabilistic foundation for the study of Bayesian regularization in deep learning, we introduce a novel probabilistic representation for the hidden layers of Convolutional Neural Networks (CNNs) \cite{CNN} by extending the Boltzmann explanation \cite{Boltzmann_machine}.
More specifically, a fully connected hidden layer formulate a multivariate discrete  Boltzmann distribution \cite{Goodfellow-et-al-2016}, and its energy function is equal to the negative activations of the layer.
In addition, a convolutional layer formulate a specific Boltzmann explanation, the Markov Random Fields (MRFs) model \cite{Geman, wang2013markov}, and its potential functions are determined by the linear convolutional filters.

Based on the proposed probabilistic representation, we demonstrate that CNNs formulate a directed acyclic graph, i.e., a Bayesian network \cite{nielsen2009bayesian, koski2011bayesian}, and SGD learns the causality between the given training dataset and training labels.
In the context of Bayesian networks, the causal links are described by conditional probabilities, thus the hidden layers close to the training dataset formulate prior distributions to describe the cause and the output layer formulates a probability to describe the expected effect.
As a result, we conclude that CNNs have explicitly Bayesian regularizations based on the correspondence between prior distributions and regularizations.

We clarify two recently observed empirical phenomena that are inconsistent with traditional theories of generalization.
First, increasing the number of hidden units can lead to a decrease in generalization error \cite{regularization5}.
Bayesian theory indicates that CNNs can use more hidden units to generate a prior distribution including more prior knowledge, thereby achieving the lower generalization error.  
Second, CNNs with good generalization performance on real labels achieve zero training error but very high generalization error on random labels \cite{generalization_regularization}.
A Bayesian network can be simplified as two components, i.e., a prior distribution and a likelihood distribution, and regularization merely corresponds to the prior distribution.
Given the same CNN architecture and the training dataset, CNNs should learn the same prior distributions to formulate the same regularization, different generalization errors result from the likelihood distribution but not related to the prior distribution, i.e., regularization.


\section{Background}

\subsection{Boltzmann distribution} 

In mathematics, the Boltzmann distribution (a.k.a, the Gibbs distribution or the renormalization group) formulates the probability of a state, e.g., $p(\boldsymbol{x}; \boldsymbol{\theta})$, as a function of the state's energy \cite{Goodfellow-et-al-2016, Boltzmann_machine, yaida2019non, energy_learning}.
\begin{equation} 
\label{Gibbs} 
{\textstyle
p(\boldsymbol{x}; \boldsymbol{\theta}) = \frac {1}{Z(\boldsymbol{\theta})}\text{exp}[-E(\boldsymbol{x; \theta})]\text{,}
}
\end{equation}
where $E(\boldsymbol{x; \theta})$ is the energy function, $Z(\boldsymbol{\theta}) = \sum_{\boldsymbol{x}}{\text{exp}[-E(\boldsymbol{x; \theta})]}$ is the partition function, and $\boldsymbol{\theta}$ denote all parameters.
A classical example of the Boltzmann distribution in machine learning is the Restricted Boltzmann Machine (RBM) \cite{deep-BM}.

The Boltzmann distribution can be reformulated as various probabilistic models through redefining the energy function $E(\boldsymbol{x; \theta})$.
In the context of probabilistic graphical model, a fundamental Boltzmann distribution is the Markov Random Fields (MRFs) and its energy function is equivalent to the summation of multiple potential functions, i.e., ${\scriptstyle E(\boldsymbol{x; \theta}) = -\sum_{k}f_k(\boldsymbol{x}; \boldsymbol{\theta}_k)}$ \cite{Geman, wang2013markov}.
Moreover, assuming a single potential function defines an expert, i.e., ${\scriptstyle \boldsymbol{F}_k = \frac {1}{Z(\boldsymbol{\theta}_k)}\text{exp}[-f_k(\boldsymbol{x}; \boldsymbol{\theta}_k)]}$, the Boltzmann distribution becomes a Product of Experts (PoE) model , i.e., ${\scriptstyle p(\boldsymbol{x}; \boldsymbol{\theta}) =  \frac {1}{Z(\boldsymbol{\theta})}\prod_{k}\boldsymbol{F}_k}$, where ${\scriptstyle Z(\boldsymbol{\theta}) = \prod_k Z(\boldsymbol{\theta}_k)}$ \cite{hinton1999products, CD}.

\subsection{Bayesian networks} 

A Bayesian network represents the causal probabilistic relationship among a set of random variables based on their conditional distributions.
 It consists of two major parts: a directed acyclic graph and a set of conditional probability distributions \cite{nielsen2009bayesian, koski2011bayesian}.
 For example, $\boldsymbol{A} \rightarrow \boldsymbol{B}$ forms a directed acyclic graph to show the possible causality between two random variables $\boldsymbol{A}$ and $\boldsymbol{B}$, which can be formulated as $P(\boldsymbol{A}, \boldsymbol{B}) = P(\boldsymbol{B|A})P(\boldsymbol{A})$, where $P(\boldsymbol{A})$ formulates a prior distribution to expresses the uncertainty of the cause and $P(\boldsymbol{B|A})$ formulates the uncertainty of the effect $\boldsymbol{B}$ given the cause $\boldsymbol{A}$.
In particular, since the prior distribution $P(\boldsymbol{A})$ expresses the prior belief of the cause that lead to the effect, $P(\boldsymbol{A})$ corresponds to the regularization in terms of the Bayesian regularization theory.

In this paper, we show that the entire architecture of CNNs forms a Bayesian network and the hidden layers close to the input formulate prior distributions to express the prior belief leading to the expected effect (i.e., labels), thus these hidden layers form explicitly Bayesian regularizations.

\section{Main results}

\subsection{Preliminaries}

We assume that $\boldsymbol{X}$ and $\boldsymbol{Y}$ are two random variables and $P_{\boldsymbol{\theta}}(\boldsymbol{X}, \boldsymbol{Y}) = P(\boldsymbol{Y|X})P(\boldsymbol{X})$ is an unknown joint distribution between $\boldsymbol{X}$ and $\boldsymbol{Y}$, where $P(\boldsymbol{X})$ describes the prior knowledge of $\boldsymbol{X}$, $P(\boldsymbol{Y|X})$ describes the statistical connection between $\boldsymbol{X}$ and $\boldsymbol{Y}$, and $\boldsymbol{\theta}$ indicate the parameters of $P_{\boldsymbol{\theta}}(\boldsymbol{X}, \boldsymbol{Y})$. 

A dataset $\boldsymbol{\mathcal{D}} = \{(\boldsymbol{x}^j, \boldsymbol{y}^j)| \boldsymbol{x}^j \in {\mathbb{R}}^{M}, \boldsymbol{y}^j \in {\mathbb{R}}^{L}\}_{j=1}^{J}$ is composed of i.i.d. samples generated from $P_{\theta}(\boldsymbol{X}, \boldsymbol{Y})$.
In addition, a CNN with $I$ hidden layers is denoted as $\text{CNN} = \{\boldsymbol{x; f_1; ...; f_I; f_Y}\}$ and trained by $\boldsymbol{\mathcal{D}}$, where $(\boldsymbol{x},\boldsymbol{y}) \in \boldsymbol{\mathcal{D}}$ are the input of the CNN and the corresponding training label, and $\boldsymbol{f_Y}$ is an estimation of the true distribution $P(\boldsymbol{Y|X})$.
As a result, $\boldsymbol{x} \sim P(\boldsymbol{X})$ and $\boldsymbol{f_Y} \approx P(\boldsymbol{Y|X})$.

\subsection{The probabilistic representation for the hidden layers of CNNs}
\label{prob_cnn_main}

In this section, we extend the Boltzmann explanations to CNNs and introduce a novel probabilistic representation for the hidden layers of CNNs.
The detailed derivation is included in Appendix \ref{prob_cnn}.
Since most CNNs only contain two categories of hidden layers, namely fully connected layers and convolutional layers with non-linear operators, e.g., ReLU and pooling, the proposed probabilistic representation can be summarized as the following propositions. 

\textbf{Proposition 1:} \textit{A fully connected layer formulates a multivariate discrete Boltzmann distribution and the energy function of which is equivalent to the negative activations of the fully connected layer.}

A fully connected layer $\boldsymbol{f_{i}} = \{f_{in}\}_{n=1}^N$ with $N$ neurons formulates a multivariate discrete Boltzmann distribution to describe the probability of $N$ features defined by the neurons occurring in the input.
Their energy functions are equal to the negative of their respective neurons, i.e., $E_{f_{in}} = -f_{in}(\boldsymbol{f_{j}})$.
\begin{equation}
\label{gibbs_layer_sim1}
P(\boldsymbol{F_i}) = \{P(f_{in}) = \frac {1}{Z_{\boldsymbol{F_i}}}\text{exp}[f_{in}(\boldsymbol{f_{j}})]\}_{n=1}^N
\end{equation}
where the previous layer $\boldsymbol{f_{j}} = \{f_{jm}\}_{m=1}^M$ is the input, $f_{in}(\boldsymbol{f_{j}}) = \sigma[g_{in}(\boldsymbol{f_j})]$ is the activation of the $n$th neuron, ${ g_{in}(\boldsymbol{f_j}) = \sum_{m=1}^M\alpha_{mn}\cdot f_{jm} + b_{in}}$ is a linear filter with $\alpha_{mn}$ being the weights and $b_{in}$ being the bias, $\sigma(\cdot)$ is an activation function, and ${ Z_{\boldsymbol{F}} = \sum_{n=1}^N\text{exp}[f_{in}(\boldsymbol{f_j})]}$ is the partition function.

\textbf{Proposition 2:} \textit{A convolutional layer formulates a specific Boltzmann distribution, i.e., the MRF model, and its energy function is equal to the negative summation of all the convolutional channels.}

A convolutional layer with $N$ convolutional channels, i.e., $\boldsymbol{f_i} = \{\boldsymbol{f^n_i} = \sigma(\varphi_i^n(\boldsymbol{f_j}))\}_{n=1}^N$, formulates a specific Boltzmann distribution, namely a MRF model.
\begin{equation}
\label{f1_mrfs}
P(\boldsymbol{F_i}) = \frac{1}{Z_{\boldsymbol{F_i}}}\text{exp}[\sum_{n=1}^N\varphi_i^n(\boldsymbol{f_j})]
\end{equation}
where $\boldsymbol{f_j}=\{\boldsymbol{f_j^q}\}_{q=1}^Q$ is the input with $Q$ channels, $\varphi_i^n(\boldsymbol{f_j}) = \sum_{q=1}^Q \boldsymbol{S^{(n, q)}} \circ \boldsymbol{f_j^q} + b_i^n \cdot \boldsymbol{1})$ is the $n$th linear convolutional channel corresponding to the potential function of a MRF model, in which $\boldsymbol{S^{(n, q)}}$ is the $n$th convolutional filter applying to the $q$th channel, $b_i^n$ is the bias,  $\sigma(\cdot)$ is an non-linear operator, and the partition function $Z_{\boldsymbol{F_i}} = \int_{\boldsymbol{f_j}}\text{exp}[\sum_{n=1}^N\varphi_i^n(\boldsymbol{f_j})]d{\boldsymbol{f_j}}$.
In particular, based on the equivalence between SGD and the first order approximation \cite{battiti1992first}, Equation \ref{f1_mrfs} indicates that the statistical property of a convolutional layer is determined by the linear convolutional operations $\{\varphi_i^n(\boldsymbol{f_j})\}_{n=1}^N$.

In summary, the Boltzmann distribution establishes a novel connection between the hidden layers of CNNs and probabilistic models through defining the energy function of a Boltzmann distribution as a function of the activations in a hidden layer. 
It is important to note that the energy function is a sufficient statistics of a Boltzmann distribution, i.e., the later is entirely dependent on the former \cite{nielsen2009statistical}.
As a result, we can specify a unique Boltzmann distribution given a hidden layer of CNNs.

Moreover, the architecture of hidden layers determines the statistical property of the corresponding Boltzmann distribution.
For example, a fully connected layer assumes all the input data being connected to each other, but a convolutional layer assumes local connectivity, i.e., the input data only depends on its neighbors. 
Therefore, their corresponding Boltzmann distributions have different statistical properties: 
Equation \ref{gibbs_layer_sim1} uses the linear filter $g_{in}(\boldsymbol{f_j})$ to describe the fully connected features, and Equation \ref{f1_mrfs} uses the potential functions of MRF to model the local connected features.

\subsection{Explicitly Bayesian regularizations in CNNs}

\textbf{Proposition 3:} \textit{The entire architecture of CNNs can be explained as a Bayesian network.}

Given a $\text{CNN} = \{\boldsymbol{x; f_1; ...; f_I; f_Y}\}$, the output of the hidden layer $\boldsymbol{f_i}$ is the input of the next hidden layer $\boldsymbol{f_{i+1}}$, 
thus the CNN forms a serial connection in terms of Bayesian networks.
\begin{equation} 
\label{mrf_dnns}
\boldsymbol{f_1 \rightarrow \cdots \rightarrow f_I \rightarrow f_Y}
\end{equation}

Since the proposed probabilistic representation specifies the distribution of the hidden layers of CNNs, the entire architecture of the CNN corresponds to a Bayesian network as
\begin{equation} 
\label{pdf_networks}
{\textstyle 
P(\boldsymbol{F_Y; F_I; \cdots; F_1|X}) = P(\boldsymbol{F_Y|F_{I}})\cdot... \cdot P(\boldsymbol{F_{i+1}|F_{i}})\cdot ... \cdot P(\boldsymbol{F_1|X})
}
\end{equation}
where $P(\boldsymbol{F_1|X})$ formulates a prior Boltzmann distribution to describe the cause, the $P(\boldsymbol{F_Y|F_I})$ formulates a Boltzmann distribution to express the expected effect, and the intermediate hidden layers formulate multiple conditional distributions to transfer the influence of the cause to the effect.

Previous works demonstrate that the output of CNNs estimates a Bayesian posterior distribution $P(\boldsymbol{F_Y|X})$ \cite{Bayesian-cnn, posterior4}.
In the context of Bayesian networks,  $P(\boldsymbol{F_Y|X})$ can be explained as a variable inference given the Bayesian networks corresponding to CNNs.
\begin{equation} 
\label{pdf_networks}
P(\boldsymbol{F_Y|X}) = \int_{\boldsymbol{F_I}} \cdots \int_{\boldsymbol{F_1}} P(\boldsymbol{F_Y; F_I..; F_1|X}) d{\boldsymbol{F_I}} \cdots d{\boldsymbol{F_1}}
\end{equation}
Since CNNs form a Bayesian network with the serial connection and the distribution of each layer is a Boltzmann distribution, we prove that $P(\boldsymbol{F_Y|X})$ is also a Boltzmann distribution and the energy function of which is also equivalent to the functionality of the entire CNN in Appendix \ref{vi_cnn}.

Given the training dataset $\boldsymbol{\mathcal{D}}$ and a predefined loss function, e.g., the cross entropy, SGD can be explained as a parameter learning procedure for $P(\boldsymbol{F_Y|X})$.
Notably, since CNNs form a serial connection indicated by Equation \ref{mrf_dnns}, both the Bayesian posterior distribution $P(\boldsymbol{F_Y|X})$ and the entire Bayesian network $P(\boldsymbol{F_Y; F_I..; F_1|X})$ correspond to the same architecture of CNNs, thus SGD can learn the parameters of the entire Bayesian network by merely optimizing $P(\boldsymbol{F_Y|X})$.

\textbf{Proposition 4:} \textit{CNNs have explicitly Bayesian regularizations} 

In the Bayesian approach, regularization is achieved by specifying a prior distribution over the parameters and subsequently averaging over the posterior distribution \cite{bayesian_regularization}. 
Therefore, the Bayesian networks explanation provides a novel perspective to clarify the generalization of CNNs. 

Above all, CNNs have explicitly Bayesian regularizations.
Since we explain the entire architecture of CNNs as a Bayesian network with the serial connection indicated by Equation \ref{mrf_dnns}, the hidden layers close to the input should formulate prior distributions, thereby representing explicit regularizations.
In contrast to previous works ascribing the generalization of CNNs to SGD implementing implicit regularization, Proposition 3 shows that SGD merely serves as a parameter learning for $P(\boldsymbol{F_Y|X})$ and explicit Bayesian regularizations are defined by the hidden layers corresponding to prior distributions.

Moreover, the Bayesian regularization explanation can clarify several recently observed empirical phenomena that are inconsistent with traditional theories of generalization \cite{generalization_regularization1}.
First, increasing the number of hidden units of CNNs, i.e., increasing the number of parameters, can lead to a decrease in generalization error \cite{regularization5}.
Bayesian theory indicates that more complex prior distributions can include more prior belief of the training dataset.
As a result, CNNs can use many hidden units to generate much powerful prior distributions for regularizing the entire network, thereby guaranteeing the generalization performance though they are over-parametrized.  

Second, CNNs with good generalization performance on real labels achieve zero training error but very high generalization error on random labels \cite{generalization_regularization}.
Bayesian theory enables us to utilize the `separate and conquer' rule to understand the special case.
In terms of Bayesian theory, a Bayesian network can be simplified as two components, namely a prior distribution and a likelihood distribution. 
Given the same CNN architecture and the same training dataset with different labels, e.g., real labels and random labels, the CNN should formulate the same regularizations, because regularization merely corresponds the prior distribution.
Different generalization errors for different labels result from the learned likelihood distribution but not related to the prior distribution.

\section{Simulations}

In this section, we use a synthetic dataset to validate explicitly Bayesian regularizations in CNNs, and demonstrate that the proposed Bayesian regularization can explain two noteworthy empirical phenomena that are inconsistent with traditional theories of generalization.
\subsection{Setup}
\label{setup}

\begin{figure}[t]
\centering
\includegraphics[scale=0.45]{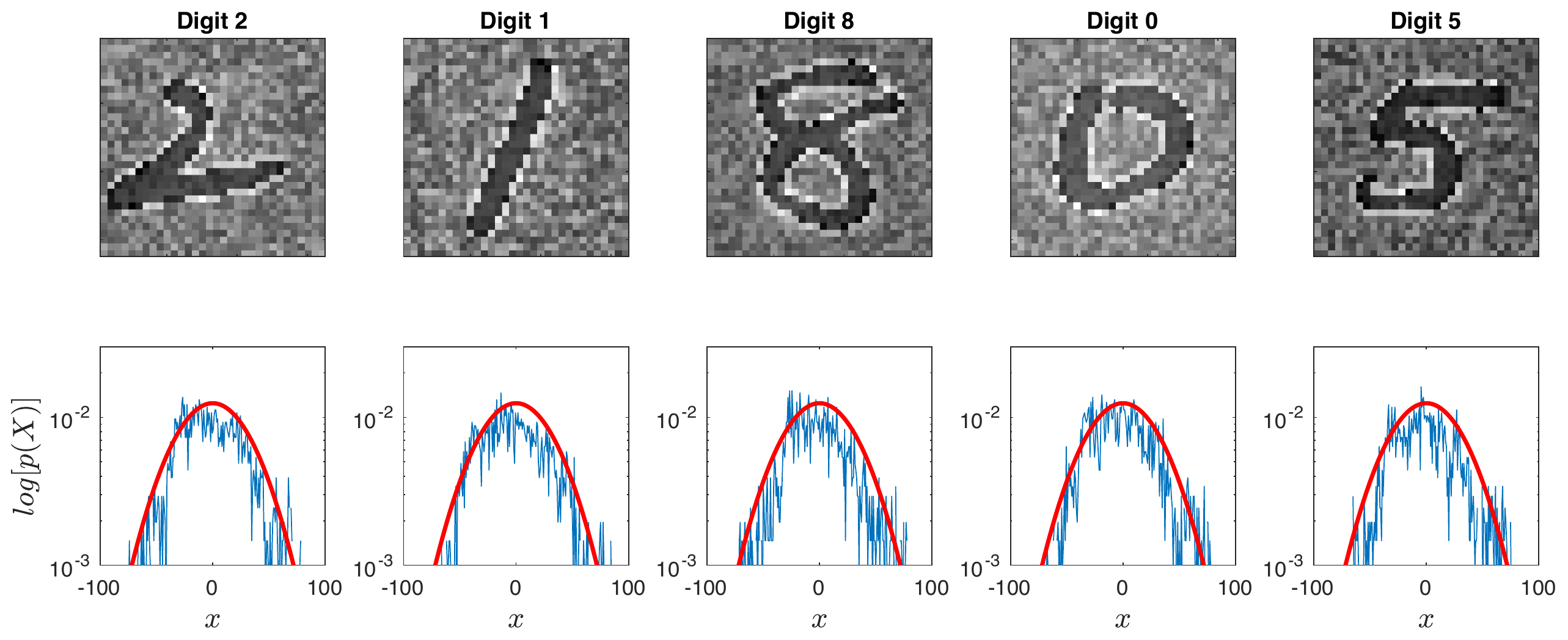}
\caption{\small{
The first row shows five synthetic images of handwritten digits, the second row shows their respective empirical distributions, and the red curve indicates the true distribution $\mathcal{N}(0, 1024)$.
}}
\label{fig_synthetic_gaussian}
\end{figure}

To validate explicitly Bayesian regularizations in CNNs, we need to firstly validate the probabilistic representation for the hidden layers of CNNs. 
However, since the distributions of benchmark datasets are unknown, it is impossible to directly use them to validate the proposed probabilistic representation.
Alternatively, we generate a synthetic dataset obeying the Gaussian distribution $\mathcal{N}(0, 1024)$ based on the NIST dataset of handwritten digits \cite{lan2019synthetic}. 
It consists of 20,000 $32 \times 32$ grayscale images in 10 classes (digits from 0 to 9), and each class has 1,000 training images and 1,000 testing images. 
The method for generating the synthetic dataset is included in Appendix \ref{syn_dataset}.

Most image datasets satisfy two fundamental assumptions: stationary and Markov \cite{stat_image}. The former means that the distribution is independent on the location, i.e., the distribution of pixels at different locations are identically distributed. The later indicates that the distribution of a single pixel is independent on the other pixels given its neighbors.
The two assumptions allow us to use an empirical distribution to simulate the true distribution of the synthetic dataset.
Figure \ref{fig_synthetic_gaussian} shows five synthetic images and their perspective empirical distributions.
It shows that their empirical distributions precisely simulate the true distribution of synthetic images, i.e., $\mathcal{N}(0, 1024)$.

We design three simple by comprehensive CNNs for classifying the synthetic dataset. 
Every CNN has two convolutional hidden layers, two max pooling layers, and a fully connected hidden layer. 
The only difference within these CNNs is that the first convolutional layer $\boldsymbol{f_1}$ of different CNNs has different number of convolutional channels.

\begin{table}[!b]
\caption{The architectures of CNNs for classifying the synthetic dataset}
\label{cnns_synthetic}
\vskip 0.15in
\begin{center}
\begin{small}
\begin{threeparttable}
\begin{tabular}{cccccc}
\toprule
R.V. & Layer & Description & CNN1 & CNN2  & CNN3 \\
\midrule
$\boldsymbol{X}$&$\boldsymbol{x}$			& Input 		& $32 \times 32 \times 1$ & $32 \times 32 \times 1$ & $32 \times 32 \times 1$ \\
\hline
\multirow{2}*{$\boldsymbol{F_1}$} &$\boldsymbol{f_1}$		& Conv ($3 \times 3$) +ReLU	& $30 \times 30 \times \boldsymbol{4}$ & $30 \times 30 \times \boldsymbol{12}$ & $30 \times 30 \times \boldsymbol{20}$ \\
&$\boldsymbol{f_2}$    	& Maxpool		& $15 \times 15 \times {4}$ & $15 \times 15 \times 12$ & $15 \times 15 \times 20$ \\
\hline
\multirow{2}*{$\boldsymbol{F_2}$}&$\boldsymbol{f_3}$    		& Conv ($5 \times 5$) + ReLU	& $11 \times 11 \times 20$ & $11 \times 11 \times 20 $ & $11 \times 11 \times 20 $ \\
&$\boldsymbol{f_4}$    	& Maxpool		& $5 \times 5 \times {20}$ & $5 \times 5 \times 20$ & $5 \times 5 \times 20 $ \\
\hline
$\boldsymbol{F_3}$&$\boldsymbol{f_5}$      	& Fully connected		& $1 \times 1 \times 20$  & $1 \times 1 \times 20$ & $1 \times 1 \times 20$ \\
\hline
$\boldsymbol{F_Y}$&$\boldsymbol{f_Y}$   	& Output(softmax)		& $1 \times 1 \times 10$  & $1 \times 1 \times 10$ & $1 \times 1 \times 10$ \\
\bottomrule
\end{tabular}
\begin{tablenotes}
            \item R.V. is the random variable of the hidden layer(s).
   \end{tablenotes}
\end{threeparttable}
\end{small}
\end{center}
\vskip -0.1in
\end{table}

Based on the proposed probabilistic representation, we identify the distribution of each hidden layer in the above $\text{CNNs} = \{\boldsymbol{x; f_1; ...; f_I; f_Y}\}$.
First, $\boldsymbol{f_1}$ formulates a specific Boltzmann distribution, namely, a MRF model.
Since the max pooling layer $\boldsymbol{f_2}$ can be viewed as an non-linear differentiable operation as ReLU, the distribution of $\boldsymbol{f_2}$ can be approximated by $\boldsymbol{f_1}$ based on the equivalence between SGD and the first order approximation \cite{battiti1992first}.
Therefore, we can use a single random variable $\boldsymbol{F_1}$ to represent the statistical properties of both $\boldsymbol{f_1}$ and $\boldsymbol{f_2}$.
Similarly, we use $\boldsymbol{F_2}$ to represent the statistical properties of both $\boldsymbol{f_3}$ and $\boldsymbol{f_4}$. 
As a result, $P(\boldsymbol{F_1})$ and $P(\boldsymbol{F_2|F_1})$ can be formulated as 
\begin{equation}
\label{cnn_f1_f2_mrfs}
\begin{split}
P(\boldsymbol{F_1}) = \frac{1}{Z_{\boldsymbol{F_1}}}\text{exp}[\sum_{n=1}^N\varphi_1^n(\boldsymbol{x})] \text{ and }
P(\boldsymbol{F_2|F_1}) = \frac{1}{Z_{\boldsymbol{F_2}}}\text{exp}[\sum_{k=1}^K\varphi_3^k(\boldsymbol{f_2})]
\end{split}
\end{equation}
where $\boldsymbol{x}$ only has a single channel since it denotes a gray scale image, and $\varphi_1^n(\boldsymbol{x}) = \boldsymbol{S_1^{n}} \circ \boldsymbol{x} + b_1^n \cdot \boldsymbol{1}$ is the $n$th linear convolutional channel in $\boldsymbol{f_1}$. 
Similarly, $\boldsymbol{f_2} = \{\boldsymbol{f^n_2}\}_{n=1}^N$ has $N$ channels, and $\varphi_3^k(\boldsymbol{f_2}) = \sum_{n=1}^N \boldsymbol{S_3^{(k, n)}} \circ \boldsymbol{f_2^k} + b_2^k \cdot \boldsymbol{1})$ is the $k$th linear convolutional channel in $\boldsymbol{f_3}$.
The two partition functions are $Z_{\boldsymbol{F_1}} = \int_{\boldsymbol{x}}\text{exp}[\sum_{n=1}^N\varphi_1^n(\boldsymbol{x})]d{\boldsymbol{x}}$ and $Z_{\boldsymbol{F_2}} = \int_{\boldsymbol{f_2}}\text{exp}[\sum_{k=1}^K\varphi_3^k(\boldsymbol{f_2})]d{\boldsymbol{f_2}}$.

Although we derive $P(\boldsymbol{F_1})$ and $P(\boldsymbol{F_2|F_1})$, it is still hard to calculate $P(\boldsymbol{F_1})$ and $P(\boldsymbol{F_2|F_1})$ because $Z_{\boldsymbol{F_1}}$ and $Z_{\boldsymbol{F_2}}$ are intractable for the high dimensional datasets $\boldsymbol{x}$ and $\boldsymbol{f_2}$.
Alternatively, since MRF models satisfy the two assumptions (i.e., stationary and Markov) and the energy function is a sufficient statistics of a Boltzmann distribution, we can use the empirical distributions of their respective energy functions (${\scriptstyle \boldsymbol{E}_{\boldsymbol{F_1}} = -\sum_{n=1}^N\varphi_1^n(\boldsymbol{x})}$ and ${\scriptstyle \boldsymbol{E}_{\boldsymbol{F_2}} = -\sum_{k=1}^K\varphi_2^k(\boldsymbol{f_2})}$) to estimate $P(\boldsymbol{F_1})$ and $P(\boldsymbol{F_2|F_1})$.
Specifically, the two empirical distributions can be expressed as
\begin{equation}
\label{cnn_f1_f2_mrfs}
\begin{split}
\hat{P}[\boldsymbol{F_1}(B_i)] = \frac{1}{C_{\boldsymbol{E}_{\boldsymbol{F_1}}}}[\sum_{nn=1}^{C_{\boldsymbol{E}_{\boldsymbol{F_1}}}} B_i \leq \boldsymbol{1}[\boldsymbol{E}_{\boldsymbol{F_1}}(nn)] \leq B_{i+1}] \\
\hat{P}[\boldsymbol{F_2|F_1}(B_i)] = \frac{1}{C_{\boldsymbol{E}_{\boldsymbol{F_2}}}}[\sum_{mm=1}^{C_{\boldsymbol{E}_{\boldsymbol{F_2}}}}B_i \leq \boldsymbol{1}[\boldsymbol{E}_{\boldsymbol{F_2}}(mm)] \leq B_{i+1}]
\end{split}
\end{equation}
where $C_{\boldsymbol{E}_{\boldsymbol{F_1}}}$ and $C_{\boldsymbol{E}_{\boldsymbol{F_2}}}$ are the number of elements in $\boldsymbol{E}_{\boldsymbol{F_1}}$ and $\boldsymbol{E}_{\boldsymbol{F_2}}$, respectively. $B_i$ and $B_{i+1}$ are the pre-defined edges, and $\boldsymbol{1}(\cdot)$ is the indicator function.

Based on the Boltzmann explanation for fully connected layers (Equation \ref{gibbs_layer_sim1}), we can formulate the distribution of $\boldsymbol{f_5}$ and $\boldsymbol{f_Y}$ as follows.
\begin{equation}
\label{fc_boltzmann}
\begin{split}
P(\boldsymbol{F_3|F_2}) = \{P(f_{5m}) &= \frac {1}{Z_{\boldsymbol{F_5}}}\text{exp}[f_{5m}(\boldsymbol{f_{4}})]\}_{m=1}^{20} \\ 
P(\boldsymbol{F_Y|F_3}) = \{P(f_{yl}) &= \frac {1}{Z_{\boldsymbol{F_Y}}}\text{exp}[f_{yl}(\boldsymbol{f_{5}})]\}_{l=1}^{10} 
\end{split}
\end{equation}
where the flattening $\boldsymbol{f_{4}}$ is the input of $P(\boldsymbol{F_3})$. In addition, the two partition functions are defined as $Z_{\boldsymbol{F_5}} = \sum_{m=1}^{20}\text{exp}[f_{5m}(\boldsymbol{f_{4}})]$ and $Z_{\boldsymbol{F_Y}} = \sum_{l=1}^{10}\text{exp}[f_{yl}(\boldsymbol{f_{5}})]$.

Based on the Bayesian networks explanation for CNNs, the above CNNs form a Bayesian network $P(\boldsymbol{F_Y; F_3; F_2; F_1}) = P(\boldsymbol{F_Y|F_{3}})\cdot P(\boldsymbol{F_{3}|F_{2}})\cdot P(\boldsymbol{F_{2}|F_{1}}) \cdot P(\boldsymbol{F_1})$,  where $P(\boldsymbol{F_1})$ formulates a prior distribution of the training dataset to describe the cause, i.e., $P(\boldsymbol{F_1}) \approx P(\boldsymbol{X})$,
$P(\boldsymbol{F_Y|F_3})$ formulates the uncertainty of the expected effect, and the intermediate layers formulate $P(\boldsymbol{F_{2}|F_{1}})$ and $P(\boldsymbol{F_{3}|F_{2}})$ to transfer the influence of the prior belief to the expected effect. 

\subsection{Validating explicitly Bayesian regularizations in CNNs}
\label{bayesian_regularization}

Since $\boldsymbol{x}$ is a sample generated from $P(\boldsymbol{X})$, i.e., $\boldsymbol{x} \sim P(\boldsymbol{X}) = \mathcal{N}(0, 1024)$, we can calculate the distance between $P(\boldsymbol{X})$ and $\hat{P}(\boldsymbol{F_1})$, i.e., $\text{KL}[P(\boldsymbol{X})||\hat{P}(\boldsymbol{F_1})]$, to examine whether $\boldsymbol{f_1}$ formulates a prior distribution of the input for validating if explicitly Bayesian regularizations exist in CNNs.

\begin{figure}[t]
\centering
\includegraphics[scale=0.5]{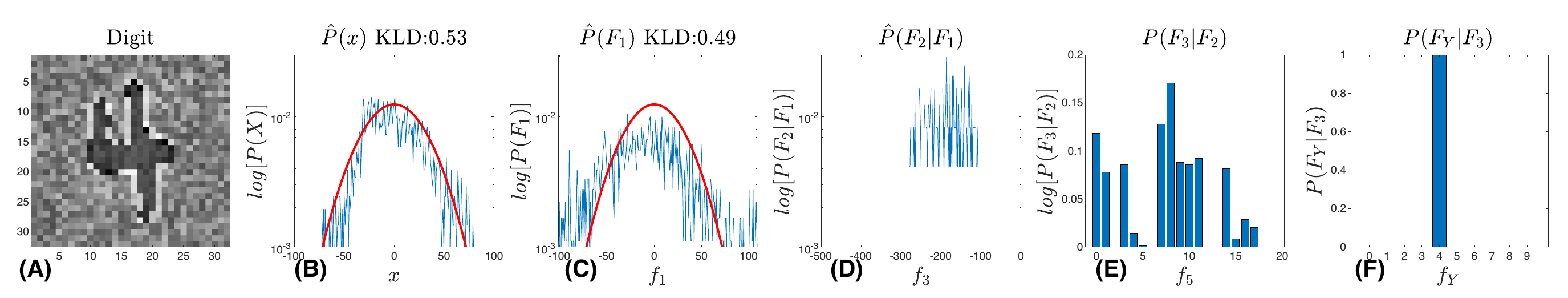}
\caption{\small{
(A) the synthetic image $\boldsymbol{x}$ is the input of CNN2,
(B) $\text{KL}[P(\boldsymbol{X})||\hat{P}(\boldsymbol{x})] = 0.53$, where the red curve indicates the truly prior distribution $P(\boldsymbol{X}) = \mathcal{N}(0, 1024)$ and $\hat{P}(\boldsymbol{x})$ is the empirical distribution of $\boldsymbol{x}$, 
(C) $\text{KL}[P(\boldsymbol{X})||\hat{P}(\boldsymbol{F_1})] = 0.49$, where $\hat{P}(\boldsymbol{F_1})$ is the empirical distribution of $\boldsymbol{f_1}$ and $\boldsymbol{f_2}$,
(D) $\hat{P}(\boldsymbol{F_2|F_1})$ is the empirical distribution of $\boldsymbol{f_3}$ and $\boldsymbol{f_4}$,
(E) the distribution ${P}(\boldsymbol{F_3|F_2})$,
and (F) the distribution ${P}(\boldsymbol{F_Y|F_3})$.
}}
\label{fig_cnn_sim1}
\end{figure}

We choose CNN2 from Table \ref{cnns_synthetic} to classify the synthetic dataset.
After CNN2 is well trained, i.e., the training error of CNN2 becomes zero, we randomly choose a synthetic image as the input of CNN2 from the testing dataset, and derive $\hat{P}(\boldsymbol{F_1})$, $\hat{P}(\boldsymbol{F_2|F_1})$, $P(\boldsymbol{F_3|F_2})$, and $P(\boldsymbol{F_Y|F_3})$. 
Finally, all the distributions are shown in Figure \ref{fig_cnn_sim1}. 
We can find that $\hat{P}(\boldsymbol{F_1})$ is indeed close to $P(\boldsymbol{X})$, because the KL divergence is samll, i.e., $\text{KL}[P(\boldsymbol{X})||\hat{P}(\boldsymbol{F_1})] = 0.49$.

This experiment validates explicitly Bayesian regularizations in CNNs.
Since we can theoretically prove CNN2 formulating a Bayesian network $P(\boldsymbol{F_Y; F_3; F_2; F_1})$ and empirically show $\hat{P}(\boldsymbol{F_1})$ modeling the prior distribution $P(\boldsymbol{X})$, CNN2 has an explicitly Bayesian regularization defined by the first convolutional layer $\boldsymbol{f_1}$.
In addition, the experiment validates the proposed probabilistic representation for the hidden layers of CNNs, especially the convolutional layer.

\subsection{The effect of increasing the number of hidden units on the generalization performance}

Based on the proposed Bayesian networks explanation and the experiment presented in Section \ref{bayesian_regularization}, we verify that $\boldsymbol{f_1}$ formulates a prior distribution as an explicit Bayesian regularization.
Since the only difference within the three CNNs is that the first convolutional layer $\boldsymbol{f_1}$ of different CNNs has different number of convolutional channels, we can use the three CNNs to examine the effect of increasing hidden units on the generalization performance of CNNs.
We use $\text{KL}[P(\boldsymbol{X})||\hat{P}(\boldsymbol{F_1})]$ to quantify the effect of increasing the number of hidden units on the prior distribution $\hat{P}(\boldsymbol{F_1})$, i.e., the Bayesian regularization.
We average $\text{KL}[P(\boldsymbol{X})||\hat{P}(\boldsymbol{F_1})]$ overall all testing images and use the testing error to quantify the generalization performance over 30 training epochs. 

\begin{figure}[b]
\centering
\includegraphics[scale=0.48]{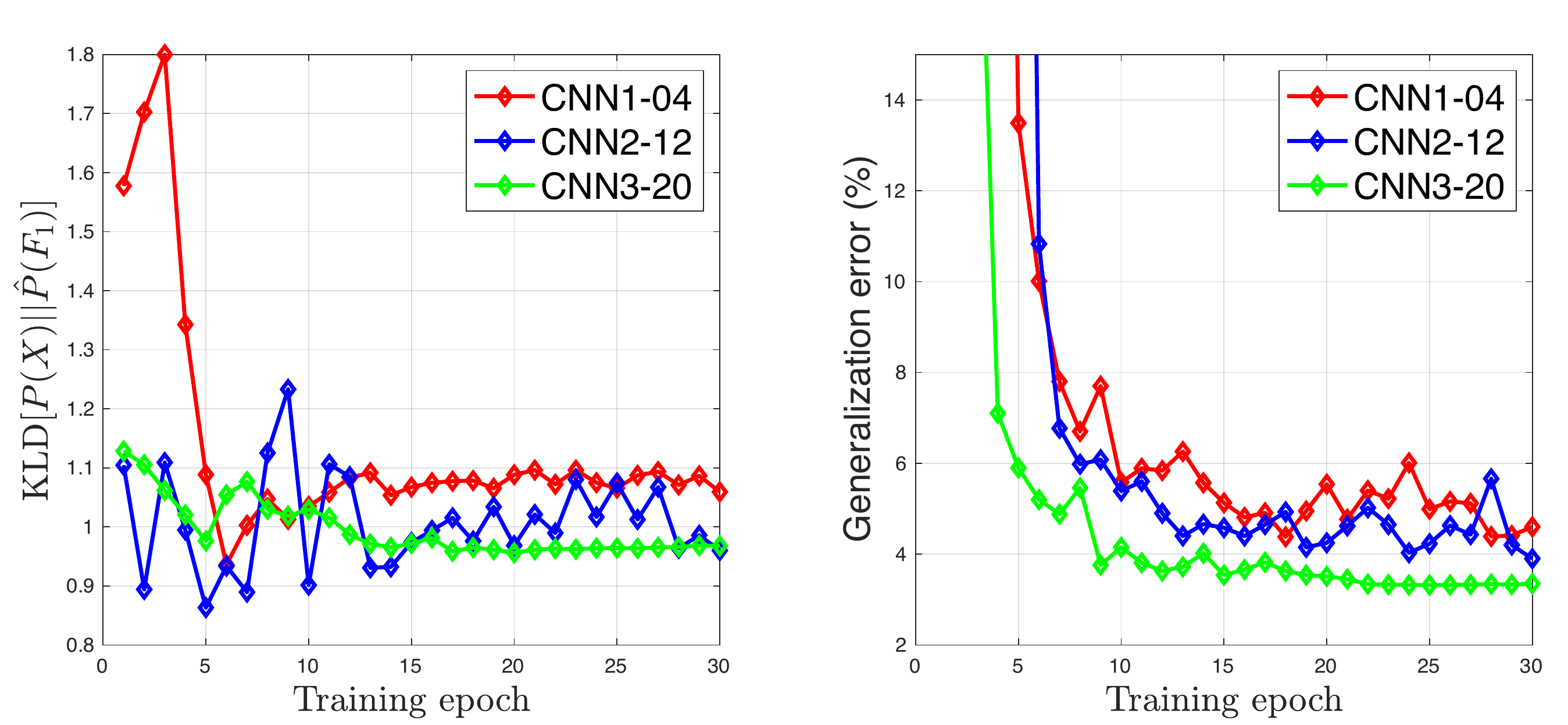}
\caption{\small{ \textbf{(Left)} 
The average $\text{KL}[P(\boldsymbol{X})||\hat{P}(\boldsymbol{F_1})]$ overall all testing images.
\textbf{(Right)}
The generalization performance of the three CNNs represented by the testing error.
The number behind the three CNNs in the legend denotes the number of hidden units of $\boldsymbol{f_1}$.
}}
\label{fig_three_cnns_generalization}
\end{figure}

Figure \ref{fig_three_cnns_generalization} shows that increasing the number of hidden units of CNNs has a negative correlation with the generalization error, which validates the Bayesian regularization explanation (Proposition 4).
Specifically, Figure \ref{fig_three_cnns_generalization} (Left) shows that $\text{KL}[P(\boldsymbol{X})||\hat{P}(\boldsymbol{F_1})]$ is decreasing as we increase the number of hidden units.
It means that $\boldsymbol{f_1}$ formulates a better Bayesian regularization as increasing the number of hidden unit.
As a result, the corresponding CNNs should achieve the lower generalization error, which is demonstrated by Figure \ref{fig_three_cnns_generalization} (Right).

\subsection{The generalization performance of CNNs on random labels}

\begin{figure}[t]
\centering
\includegraphics[scale=0.5]{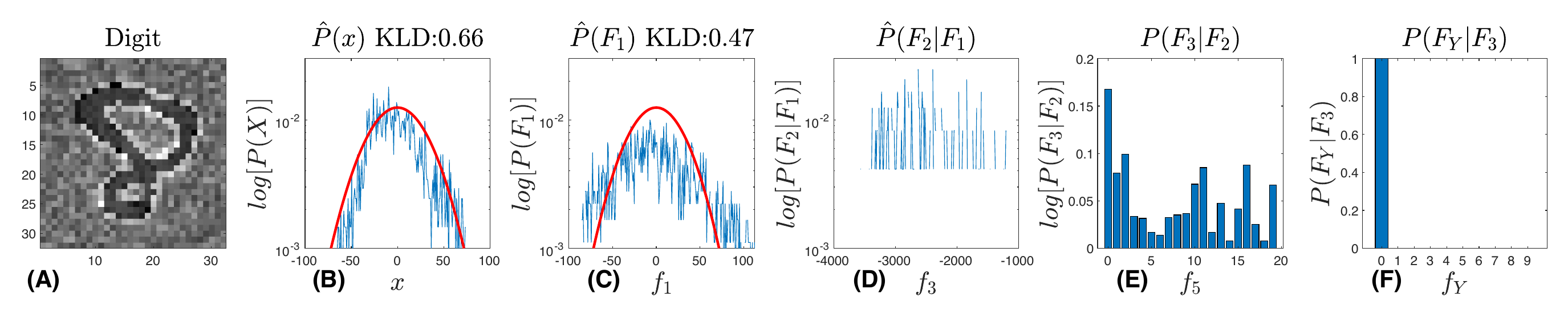}
\caption{\small{
The distribution of the hidden layers in CNN3 given a synthetic testing image with the random label. 
The notation is the same as Figure \ref{fig_cnn_sim1}, e.g., $\text{KL}[P(\boldsymbol{X})||\hat{P}(\boldsymbol{x})] = 0.66$, and $\text{KL}[P(\boldsymbol{X})||\hat{P}(\boldsymbol{F_1})] = 0.47$.
}}
\label{fig_distribution_random}
\end{figure}

\begin{figure}[htp]
\centering
\includegraphics[scale=0.25]{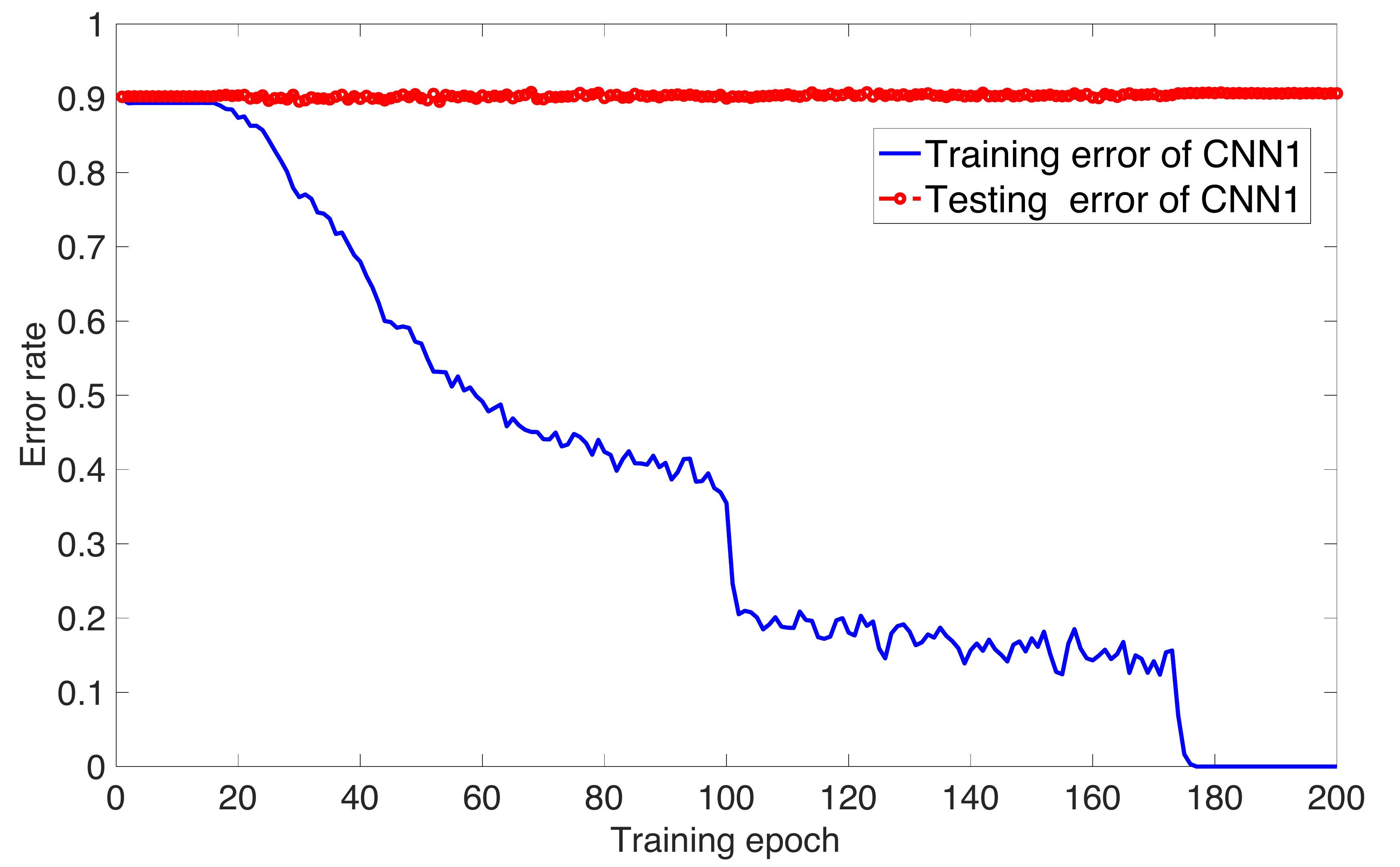}
\caption{\small{
The training and testing errors of CNN3 on random labels over 200 training epochs.
}}
\label{fig_cnn_random}
\end{figure}

Similar to the notable experiment presented in \cite{generalization_regularization}, we use CNN3 to classify the synthetic dataset with random labels. 
Figure \ref{fig_cnn_random} shows CNN3 achieving zero training error but very high testing error.
We visualize the distribution of hidden layers in CNN3 given a testing image with a random label, and Figure \ref{fig_distribution_random}(C) shows that CNN3 still can learn an accurate prior distribution $\hat{P}(\boldsymbol{F_1})$.
In this sense, CNN3 still achieves good generalization performance on random labels.

We conjecture that CNNs cannot classify random labels.
In terms of Bayesian networks, CNNs describe the causality between the given dataset and the corresponding labels, which can be simplified as the joint distribution $P(\boldsymbol{X}, \boldsymbol{Y}) = P(\boldsymbol{Y|X})P(\boldsymbol{X})$, where $P(\boldsymbol{X})$ corresponds to the prior belief, and $P(\boldsymbol{Y|X})$ describes the statistical causality.
Leveraging their powerful representation ability, CNNs definitely can learn a $P(\boldsymbol{Y|X})$ from the training dataset and corresponding random labels, thus the training error is zero.
However, the causality represented by the learned $P(\boldsymbol{Y|X})$ is obviously not consistent with the testing labels, because labels are random, thus the testing error is always around $90\%$.
In summary, the high testing error on random labels is because $P(\boldsymbol{Y|X})$ cannot model two independent variables, but not related to the prior distribution, i.e., Bayesian regularization.

\section{Conclusion}

In this paper, we demonstrate CNNs have explicitly Bayesian regularizations. 
First, we introduce a novel probabilistic representation for the hidden layers of CNNs and demonstrate that CNNs correspond to a Bayesian network with the serial connection.
We conclude that CNNs have explicitly Bayesian regularizations because the hidden layers close to the input formulate prior distributions.
Moreover, based on the proposed Bayesian regularization, we clarify two recently observed empirical phenomena that are inconsistent with traditional theories of generalization.
In the future, we hope to extend the proposed Bayesian regularization theory to general DNNs, e.g., the residual networks \cite{Deep_resiudal}, and to design new regularization algorithms through pre-training the hidden layers corresponding to prior distributions for improving the generalization performance of DNNs.

\bibliography{newinml_2019}

\clearpage

\section{Appendix}

\subsection{Proof: the proposed probabilisitic representation for the hidden layers of CNNs}
\label{prob_cnn}

\begin{figure}[t]
\centering
\includegraphics[scale=0.16]{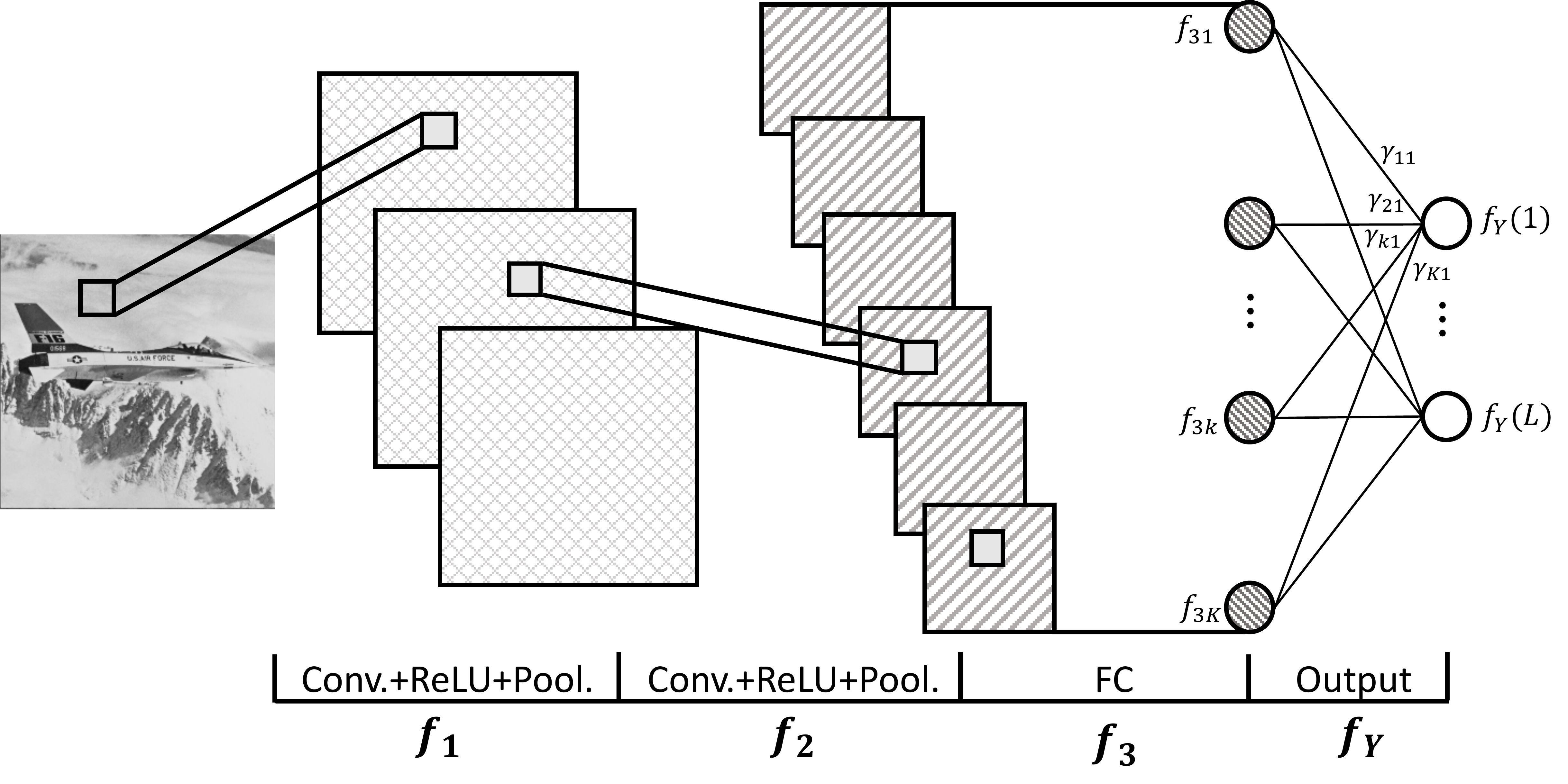}
\caption{\small{
The $\text{CNN} = \{\boldsymbol{x; f_1; f_2; f_3; f_Y}\}$ has two convolutional hidden layers $\boldsymbol{f_1}$ and $\boldsymbol{f_2}$ and a fully connected hidden layer $\boldsymbol{f_3}$.
The flattening $\boldsymbol{f_2}$ is the input of $\boldsymbol{f_3}$.
}}
\label{Img_cnn_prob}
\end{figure}

Based on the CNN show in Figure \ref{Img_cnn_prob}, we prove the proposed probabilistic representation for the hidden layers of CNNs in Section \ref{prob_cnn_main}.
In the above CNN, the input $\boldsymbol{x}$ has $M$ channels $\boldsymbol{x}=\{\boldsymbol{x}^{m}\}_{m=1}^M$ (e.g., $M=3$ if $\boldsymbol{x}$ are color images), $\boldsymbol{f_1}$ has $N$ convolutional channels, i.e., $\boldsymbol{f_1} = \{\boldsymbol{f_1^n}\}_{n=1}^N$, 
and $\boldsymbol{f_2}$ has $Q$ convolutional channels, i.e., $\boldsymbol{f_2} = \{\boldsymbol{f_2^q}\}_{q=1}^Q$.
In addition, $\boldsymbol{f_3}$ has $K$ neurons, i.e., $\boldsymbol{f_3} = \{\boldsymbol{f_{3k}}\}_{k=1}^K$.

Assuming the output layer $\boldsymbol{f_Y}$ is the softmax with $L$ nodes, its distribution can be formulated as
\begin{equation}
\label{Gibbs_fY}
P(\boldsymbol{F_Y}) = \{P(f_{yl}) = \frac {1}{Z_{\boldsymbol{F_Y}}}\text{exp}(f_{yl})\}_{l=1}^L
\end{equation}
where $Z_{\boldsymbol{F_Y}} = \sum_{l=1}^L\text{exp}(f_{yl})$ is the partition function.
Since $f_{yl} = \sum_{k=1}^{K}\gamma_{kl} \cdot f_{3k} + b_{yl}$, we have
\begin{equation} 
P(\boldsymbol{F_Y}) = \{P(f_{yl}) = \frac {1}{Z_{\boldsymbol{Y}}}\text{exp}(\sum_{k=1}^{K}\gamma_{kl} \cdot f_{3k} + b_{yl})\}_{l=1}^L
\end{equation}
Based on the properties of the exponential function, i.e., ${\textstyle \text{exp}(a+b) = \text{exp}(a)\cdot \text{exp}(b)}$ and ${\textstyle \text{exp}(a\cdot b) = [\text{exp}(b)]^{a}}$, we can reformulate $P(\boldsymbol{F_Y})$ as 
\begin{equation} 
P(\boldsymbol{F_Y}) = \{P(f_{yl}) = \frac {1}{Z'_{\boldsymbol{Y}}}\prod_{k=1}^K[\text{exp}(f_{3k})]^{\gamma_{kl}}\}_{l=1}^L
\end{equation}
where $Z'_{\boldsymbol{Y}} = Z_{\boldsymbol{Y}}/\text{exp}(b_{yl})$. Since $\{\text{exp}(f_{3k})\}_{k=1}^K$ are scalar, we can introduce a new partition function ${\textstyle Z_{\boldsymbol{F_3}} = \sum_{k=1}^K\text{exp}(f_{3k})}$ such that $\{\frac{1}{Z_{\boldsymbol{F_3}}}\text{exp}(f_{3k})\}_{k=1}^K$ becomes a probability measure.
As a result, we can further reformulate $P(\boldsymbol{F_Y}) = \{P(f_{yl})\}_{l=1}^L$ as a Product of Expert (PoE) model as
\begin{equation} 
\label{FoE_fY}
P(\boldsymbol{F_Y}) = \{P(f_{yl})= \frac {1}{Z''_{\boldsymbol{Y}}}\prod_{k=1}^K[\frac{1}{Z_{\boldsymbol{F_3}}}\text{exp}(f_{3k})]^{\gamma_{kl}}\}_{l=1}^L
\end{equation}
where ${\textstyle Z''_{\boldsymbol{Y}} = Z_{\boldsymbol{Y}}/[\text{exp}(b_{yl}) \cdot \prod_{k=1}^K[Z_{\boldsymbol{F_3}}]^{\gamma_{kl}}}]$ and each expert is defined as $\frac{1}{Z_{\boldsymbol{F_3}}}\text{exp}(f_{3k})$.

It is noteworthy that all the experts $\{\frac{1}{Z_{\boldsymbol{F_3}}}\text{exp}(f_{3k})\}_{k=1}^K$ form a probability measure and establish an exact one-to-one correspondence to all the neurons in the third hidden layer $\boldsymbol{f_3}$, i.e., $\{f_{3l}\}_{k=1}^K$.
Therefore, the distribution of $\boldsymbol{f_3}$ can be expressed as
\begin{equation}
\label{Gibbs_f3}
P(\boldsymbol{F_3}) = \{P(f_{3k}) = \frac {1}{Z_{\boldsymbol{F_3}}}\text{exp}(f_{3k})\}_{k=1}^K
\end{equation}

Based on the definition of Boltzmann distribution, Equation \ref{Gibbs_fY} and \ref{Gibbs_f3} show that $\boldsymbol{f_Y}$ and $\boldsymbol{f_3}$ formulate two multivariate Boltzmann distributions and their energy functions are equivalent to the negative of the output nodes $\{f_{yl}\}_{l=1}^L$ and the neurons $\{f_{3k}\}_{k=1}^K$, respectively.

In summary, the above derivation proves Proposition 1, i.e., a fully connected layer formulates a multivariate discrete Boltzmann distribution and the energy function of which is equivalent to the negative activations of the fully connected layer. 
The subsequent derivation proves Proposition 2.

Since $\{f_{3k} = \sum_{j=1}^J\beta_{jk} \cdot f_{2j} + b_{3k}\}_{k=1}^K$, we can derive
\begin{equation}
P(\boldsymbol{F_3}) = \{P(f_{3k}) = \frac {1}{Z_{\boldsymbol{F_3}}}\text{exp}(\sum_{j=1}^J\beta_{jk} \cdot f_{2j} + b_{3k})\}_{k=1}^K
\end{equation}
where $\{f_{2j}\}_{j=1}^J$ is the flattened output of $\boldsymbol{f_2}$, $\beta_{jk}$ is the weight of the edge between $f_{2j}$ and $f_{3k}$, and $b_{3k}$ denotes the bias. 
Since $\boldsymbol{f_2}$ is a convolutional layer with $Q$ convolutional channels, it can be reformulated as $\{f_{2j}\}_{j=1}^J = \{\boldsymbol{f^q_{2}}\}_{q=1}^Q$. 
As a result, $P(\boldsymbol{F_3})$ can be reformulated as
\begin{equation}
P(\boldsymbol{F_3}) = \{P(f_{3k}) = \frac {1}{Z_{\boldsymbol{F_3}}}\text{exp}(\sum_{q=1}^Q\boldsymbol{\beta}^{T}_{qk} \cdot \boldsymbol{f^q_{2}} + b_{3k})\}_{k=1}^K
\end{equation}
where $\boldsymbol{\beta}_{qk}$ is a subset of the parameters $\{\beta_{jk}\}_{j=1}^J$ such that $\sum_{q=1}^Q\boldsymbol{\beta}^{T}_{qk}\cdot \boldsymbol{f}^q_{2} = \sum_{j=1}^J\beta_{jk} \cdot f_{2j}$, and $\boldsymbol{\beta}^T_{qk}$ is the transpose of $\boldsymbol{\beta}_{qk}$.
Therefore, $P(\boldsymbol{F_3})$ can be reformulated as 
\begin{equation}
P(\boldsymbol{F_3}) = \{P(f_{3k}) = \frac {1}{Z'_{\boldsymbol{F_3}}}\prod_{q=1}^Q\text{exp}(\boldsymbol{\beta}^{T}_{qk}\cdot \boldsymbol{f^q_{2}})\}_{k=1}^K
\end{equation}
where $Z'_{\boldsymbol{F_3}} = Z_{\boldsymbol{F_3}}/[\text{exp}(b_{3k})]$.
Recall the element-wise matrix power, e.g., $\text{exp}(\boldsymbol{a})^{\boldsymbol{b}}$, where $\boldsymbol{a} = [a_1; a_2; a_3]$ and $\boldsymbol{b} = [b_1; b_2; b_3]$, we can derive that $\text{exp}(\boldsymbol{a}) = [\text{exp}(a_1); \text{exp}(a_2); \text{exp}(a_3)]$ and $\text{exp}(\boldsymbol{a})^{\boldsymbol{b}} = [\text{exp}(a_1)^{b_1}; \text{exp}(a_2)^{b_2}; \text{exp}(a_3)^{b_3}] = [\text{exp}(a_1 b_1); \text{exp}(a_2 b_2); \text{exp}(a_3 b_3)]$.
As a result, $\text{exp}(\boldsymbol{ab}) = \text{exp}(a_1 b_1 + a_2 b_2 + a_3 b_3) = \prod_{|a|}\text{exp}(\boldsymbol{a})^{\boldsymbol{b}}$, where $|\boldsymbol{a}|$ is the number of elements in $\boldsymbol{a}$.

Based on the element-wise matrix power, we can reformulate $P(\boldsymbol{F_3})$ as 
\begin{equation}
P(\boldsymbol{F_3}) = \{P(f_{3k}) \approx \frac {1}{Z'_{\boldsymbol{F_3}}}\prod_{q=1}^Q\text{exp}(\boldsymbol{\beta}^{T}_{qk}\cdot \boldsymbol{f^q_{2}}) = \frac {1}{Z'_{\boldsymbol{F_3}}}\prod_{q=1}^Q\prod_{|\boldsymbol{f^q_{2}}|}[\text{exp}(\boldsymbol{f^q_{2}})]^{\boldsymbol{\beta}_{qk}}\}_{k=1}^K
\end{equation}

Moreover, we introduce a new partition function $Z_{\boldsymbol{F^q_2}} = \sum_{\boldsymbol{f^q_2}}\text{exp}(\boldsymbol{f^q_{2}})$ to guarantee that $\frac{1}{Z_{\boldsymbol{F^q_2}}}\text{exp}(\boldsymbol{f^q_{2}})$ is a probability measure. As a result, $P(\boldsymbol{F_3})$ can be reformulated as
\begin{equation}
P(\boldsymbol{F_3}) = \{P(f_{3k}) = \frac {1}{Z''_{\boldsymbol{F_3}}}\prod_{q=1}^Q\prod_{|\boldsymbol{f^q_2}|}[\frac{1}{Z_{\boldsymbol{F^q_2}}}\text{exp}(\boldsymbol{f^q_{2}})]^{\boldsymbol{\beta}_{qk}}\}_{k=1}^K
\end{equation}
where $Z''_{\boldsymbol{F_3}} = Z'_{\boldsymbol{F_3}}/\prod_{q=1}^Q[Z_{\boldsymbol{F^q_2}}]^{\boldsymbol{\beta}_{qk} \cdot |\boldsymbol{f^q_2}|}$.
Overall, $P(\boldsymbol{F_3})$ can be reformulated as a PoE model, in which each expert is defined by every convolutional channel in $\boldsymbol{f_2}$
\begin{equation}
\label{f2_PoE}
P(\boldsymbol{f}^q_{2}) = \frac{1}{Z_{\boldsymbol{F^q_2}}}\text{exp}(\boldsymbol{f^q_{2}})
\end{equation}
Since $\boldsymbol{f_2}$ has $K$ convolutional channels, i.e., $\boldsymbol{f_2} = \{\boldsymbol{f^q_{2}}\}_{q=1}^Q$, and each channel can be formulated as the summation of all the convolutional channels in $\boldsymbol{f_1}$, i.e., 
\begin{equation}
\label{f2_conv}
\boldsymbol{f_{2}} = \{\boldsymbol{f^q_{2}} = \sigma_2(\sum_{n=1}^N \boldsymbol{S^{(q, n)}_{2}} \circ \boldsymbol{f^n_1} + b^q_{2}\cdot \boldsymbol{1})\}_{q=1}^Q
\end{equation} 
where $\boldsymbol{S^{(q, n)}_{2}} \circ \boldsymbol{f^n_1}$ is the output of the $q$th convolutional filter applying into the $n$th channel of $\boldsymbol{f_1}$, $b_2^q$ is the bias, and $\sigma_2(\cdot)$ is the activation function. Therefore, we have
\begin{equation}
\label{f2_PoE1}
P(\boldsymbol{f}^q_{2}) = \frac{1}{Z_{\boldsymbol{F^q_2}}}\text{exp}[\sigma_2(\sum_{n=1}^N \boldsymbol{S^{(q, n)}_{2}} \circ \boldsymbol{f^n_1} + b^q_{2}\cdot \boldsymbol{1})]
\end{equation}

Based on the equivalence between the gradient descent learning and the first order approximation (Appendix \ref{gd_first_order}), Equation \ref{f2_PoE1} can be approximated as
\begin{equation}
\label{f2_PoE2}
P(\boldsymbol{f}^q_{2}) \approx \frac{1}{Z_{\boldsymbol{F^q_2}}}\text{exp}(\sum_{n=1}^N \boldsymbol{S^{(q, n)}_{2}} \circ \boldsymbol{f^n_1} + b^q_{2}\cdot \boldsymbol{1})
\end{equation}

Let us define a linear function as $\varphi^q_2(\boldsymbol{f_{1}}) = \sum_{n=1}^N \boldsymbol{S^{(q, n)}_{2}} \circ \boldsymbol{f^n_1} + b^q_{2}\cdot \boldsymbol{1}$, thus the distribution of the $q$th convolutional channel in $\boldsymbol{f_2}$, i.e., $P(\boldsymbol{f}^q_{2})$, can be expressed as
\begin{equation}
\label{f2_PoE3}
P(\boldsymbol{f}^q_{2}) \approx \frac{1}{Z_{\boldsymbol{F^q_2}}}\text{exp}[\varphi^q_2(\boldsymbol{f_{1}})]
\end{equation}

If we regard all the linear filters $\{\varphi^q_2(\boldsymbol{f_{1}})\}_{k=1}^K$ as potential functions for modeling signal structures of $\boldsymbol{f_1}$, we can formulate $P(\boldsymbol{F_2})$ as a specific Boltzmann distribution, i.e., the Markov Random Fields (MRFs), which can be expressed as
\begin{equation}
\label{f2_mrfs}
P(\boldsymbol{F_2}) = \frac{1}{Z_{\boldsymbol{F_2}}}\prod_{q=1}^QP(\boldsymbol{f}^q_{2}) = \frac{1}{Z_{\boldsymbol{F_2}}}\text{exp}[\sum_{q=1}^Q\varphi^q_2(\boldsymbol{f_{1}})]
\end{equation}
where $Z_{\boldsymbol{F_2}} = \int_{\boldsymbol{f_1}}\text{exp}[\sum_{q=1}^Q\varphi^k_2(\boldsymbol{f_{1}})]d\boldsymbol{f_1}$ is the partition function for $P(\boldsymbol{F_2})$.
In summary, we prove that the convolutional layer $\boldsymbol{f_2}$ formulates a MRF model to describe the statistical property of $\boldsymbol{f_1}$. 

\begin{figure}[!t]
\centering
\includegraphics[scale=0.23]{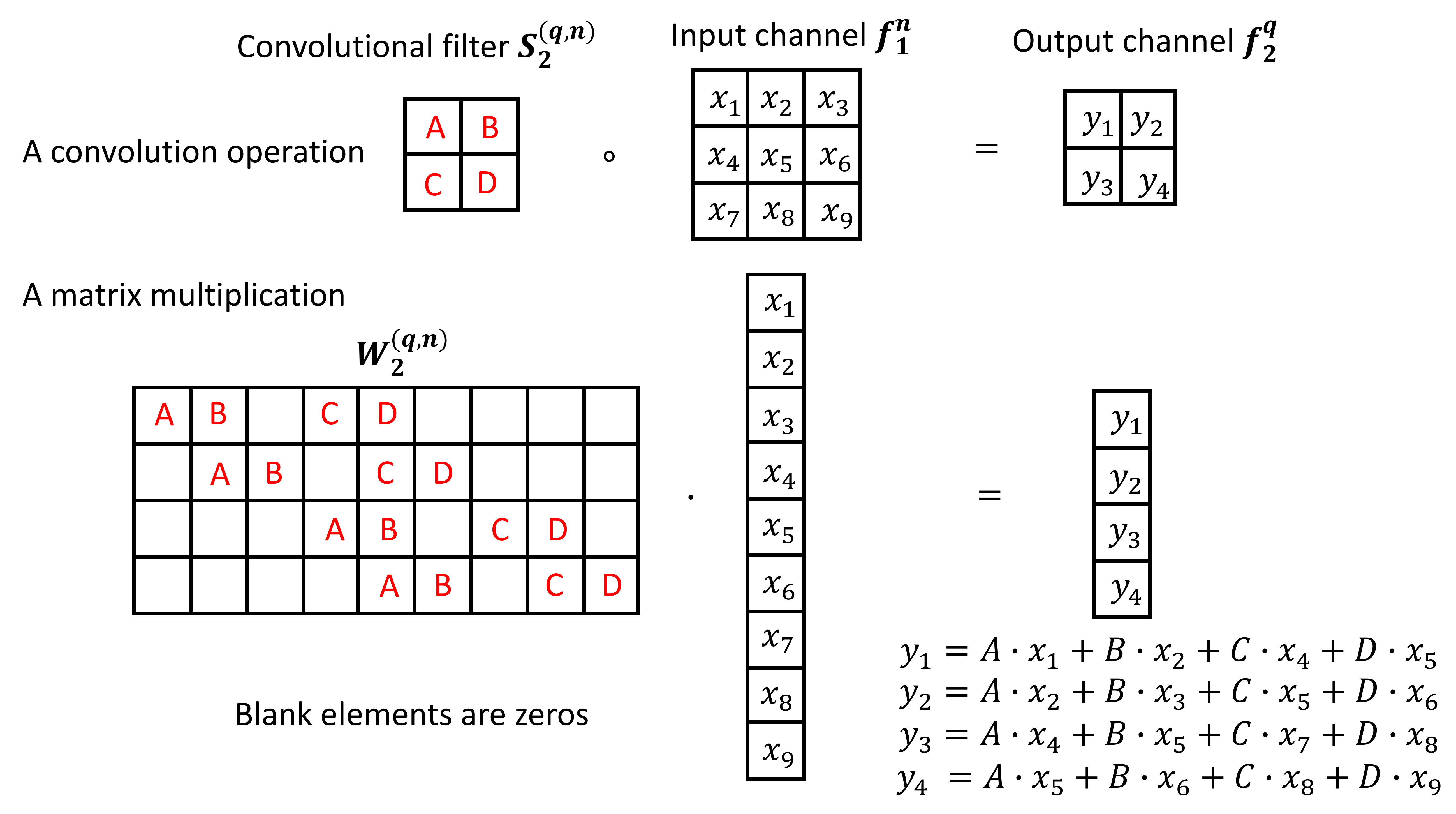}
\caption{\small{
The equivalence between a convolutional operation and a matrix multiplication. 
}}
\label{Img_matrix_mulp2}
\end{figure}

Subsequently, we prove that the convolutional layer $\boldsymbol{f_1}$ also formulates a MRF model to describe the statistical property of the input $\boldsymbol{x}$.
Based on Equation \ref{f2_PoE2}, we have
\begin{equation}
P(\boldsymbol{f}^q_{2}) \approx \frac{1}{Z'_{\boldsymbol{F^q_2}}}\text{exp}(\sum_{n=1}^N \boldsymbol{S^{(q, n)}_{2}} \circ \boldsymbol{f^n_1})
\end{equation}
where $Z'_{\boldsymbol{F^k_2}} = Z_{\boldsymbol{F^q_2}}/\text{exp}(b^q_2)$. 
Based on the equivalence between a convolutional channel and a matrix multiplication proven in Appendix \ref{necessary_conv}, we can regard a convolutional operation as a matrix multiplication, which is visualized in Figure \ref{Img_matrix_mulp2}.
As a result, we have
\begin{equation}
P(\boldsymbol{f}^q_{2}) \approx \frac{1}{Z'_{\boldsymbol{F^q_2}}}\text{exp}(\sum_{n=1}^N \boldsymbol{S^{(q, n)}_{2}} \circ \boldsymbol{f^n_1}) = \frac{1}{Z'_{\boldsymbol{F^q_2}}}\text{exp}(\sum_{n=1}^N\boldsymbol{W^{(q, n)}_{2}} \cdot \boldsymbol{f^n_1})
\end{equation}
where $\boldsymbol{W^{(q, n)}_{2}}$ is a matrix corresponding to the convolution filter $\boldsymbol{S^{(q, n)}_{2}}$.

Based on the element-wise matrix power, $P(\boldsymbol{f}^q_{2})$ can be reformulated as
\begin{equation}
P(\boldsymbol{f}^q_{2}) \approx \frac{1}{Z'_{\boldsymbol{F^q_2}}}\prod_{n=1}^N\text{exp}(\boldsymbol{W^{(q, n)}_{2}} \cdot \boldsymbol{f^n_1}) = \frac{1}{Z'_{\boldsymbol{F^q_2}}}\prod_{n=1}^N\text{exp}(\boldsymbol{f^n_1})^{\boldsymbol{W^{(q, n)}_{2}}}
\end{equation}

Similarly, we introduce a new partition function $Z_{\boldsymbol{F^n_1}} = \int_{\boldsymbol{f^n_1}}\text{exp}(\boldsymbol{f^n_{1}})d\boldsymbol{f^n_1}$ to guarantee that $\frac{1}{Z_{\boldsymbol{F^n_1}}}\text{exp}(\boldsymbol{f^n_{1}})$ is a probability measure. As a result, $P(\boldsymbol{f^q_2})$ can be reformulated as
\begin{equation}
P(\boldsymbol{f^q_2}) = P(f_{yl}) \approx \frac {1}{Z''_{\boldsymbol{F^q_2}}}\prod_{n=1}^N[\frac{1}{Z_{\boldsymbol{F^n_1}}}\text{exp}(\boldsymbol{f^n_{1}})]^{\boldsymbol{W^{(q,n)}_{2}}}
\end{equation}
where $Z''_{\boldsymbol{F^q_2}} = Z'_{\boldsymbol{F^q_2}}/\prod_{n=1}^N[Z_{\boldsymbol{F^n_1}}]^{\boldsymbol{W^{(q,n)}_{2}}}$.
Therefore, $P(\boldsymbol{f^q_2})$ can be reformulated as a PoE model, in which all the experts are defined as $P(\boldsymbol{f^n_1}) = \frac{1}{Z_{\boldsymbol{F^n_1}}}\text{exp}(\boldsymbol{f^n_{1}})$.

Since $\boldsymbol{f_1}$ has $N$ convolutional channels
\begin{equation}
\label{f2_conv}
\boldsymbol{f_1} = \{\boldsymbol{f^n_{1}} = \sigma_1(\sum_{m=1}^M \boldsymbol{S^{(n, m)}_{1}} \circ \boldsymbol{x^m} + b^n_{1}\cdot \boldsymbol{1})\}_{n=1}^N
\end{equation} 

Based on the same derivation as Equations \ref{f2_PoE2}, \ref{f2_PoE3}, \ref{f2_mrfs}, $P(\boldsymbol{F_1})$ also can be expressed as a MRF model 
\begin{equation}
P(\boldsymbol{F_1}) = \frac{1}{Z_{\boldsymbol{F_1}}}\prod_{n=1}^NP(\boldsymbol{f}^n_{1}) = \frac{1}{Z_{\boldsymbol{F_1}}}\text{exp}[\sum_{n=1}^N\varphi^n_1(\boldsymbol{x})]
\end{equation}
where $\varphi^n_1(\boldsymbol{x}) = \sum_{m=1}^M \boldsymbol{S^{(n, m)}_{1}} \circ \boldsymbol{x^m} + b^n_{1}\cdot \boldsymbol{1})$ and $Z_{\boldsymbol{F_1}} = \int_{\boldsymbol{x}}\text{exp}[\sum_{n=1}^N\varphi^n_1(\boldsymbol{x})]d\boldsymbol{x}$.
Overall, the distribution of a convolutional layer can be formulated as a MRF model.

\subsection{Proof: the variable inference of the Bayesian networks corresponding to the CNNs}
\label{vi_cnn}

Given a $\text{CNN} = \{\boldsymbol{x; f_1; ...; f_I; f_Y}\}$, we demonstrate the entire architecture of the CNN corresponds to a Bayesian network as
\begin{equation} 
\label{pdf_networks}
{\textstyle 
P(\boldsymbol{F_Y; F_I; \cdots; F_1|X}) = P(\boldsymbol{F_Y|F_{I}})\cdot... \cdot P(\boldsymbol{F_{i+1}|F_{i}})\cdot ... \cdot P(\boldsymbol{F_1|X})
}
\end{equation}
Here we prove that the posterior distribution $P(\boldsymbol{F_Y|X})$ is a Boltzmann distribution, and the energy function of which is also equivalent to the functionality of the entire CNN.

Based on the variable elimination, $P(\boldsymbol{F_Y|X})$ can be formulated as
\begin{equation} 
\label{pdf_networks}
\begin{split}
P(\boldsymbol{F_Y|X}) &= \int_{\boldsymbol{F_I}} \cdots \int_{\boldsymbol{F_1}} P(\boldsymbol{F_Y; F_I..; F_1|X}) d{\boldsymbol{F_I}} \cdots d{\boldsymbol{F_1}} \\
&= P(\boldsymbol{F_Y|F_{I}}) \int_{\boldsymbol{F_I}} P(\boldsymbol{F_I|F_{I-1}})d{\boldsymbol{F_I}} \cdots \int_{\boldsymbol{F_{i}}} P(\boldsymbol{F_{i}|F_{i-1}})d{\boldsymbol{F_i}} \cdots \int_{\boldsymbol{F_1}} P(\boldsymbol{F_1|X}) d{\boldsymbol{F_1}}
\end{split}
\end{equation}
where $P(\boldsymbol{F_{i}|F_{i-1}}) = \frac{1}{\boldsymbol{Z_{F_{i}}}}\text{exp}[-E_{\boldsymbol{F_{i}}}(\boldsymbol{f_{i-1}})]$ is a Boltzmann distribution.
All the above integrations, e.g., $\int_{\boldsymbol{F_{i}}} P(\boldsymbol{F_{i}|F_{i-1}})d{\boldsymbol{F_i}}$, are equal to 1, because the integral space, i.e., $\boldsymbol{F_{i}}$, are entirely depends on $\boldsymbol{f_{i-1}}$. As a result, $P(\boldsymbol{F_Y|X}) = P(\boldsymbol{F_Y|F_{I}})$.
Since $\boldsymbol{f_Y}$ formulates a Boltzmann distribution, and its energy function is $E_{\boldsymbol{F_Y}}(\boldsymbol{f_I}) = -f_Y(f_I(f_{I-1}\cdots f_1(\boldsymbol{x})))$, we have 
\begin{equation} 
{\textstyle 
P(\boldsymbol{F_Y|X}) = \frac{1}{\boldsymbol{Z_{F_{Y}}}}\text{exp}[-f_Y(f_I(f_{I-1}\cdots f_1(\boldsymbol{x})))]
}
\end{equation}
Therefore $P(\boldsymbol{F_Y|X})$ is also a Boltzmann distribution, and the energy function of which is also equivalent to the functionality of the entire CNN.

\clearpage

\subsection{The method for generating the synthetic dataset}
\label{syn_dataset}

\begin{algorithm}[!t]  
  \caption{The algorithm for generating the synthetic dataset}  
  \label{bcnn_infer}  
  \begin{algorithmic}[1]  
    \Require
     NIST dataset of handwritten digits by class
    \begingroup
    \everymath{\footnotesize}
    \small  
    \Repeat      
      \State binarizing an image of NIST to obtain $\boldsymbol{z}$
      \State extracting the central part of $\boldsymbol{z}$ to obtain $\boldsymbol{z}_c$ with dimension $64 \times 64$
      \State downsampling $\boldsymbol{z}_c$ to obtain $\boldsymbol{z}_{cd}$ with dimension $32 \times 32$
      \State extracting the edge of $\boldsymbol{z}_{cd}$ to obtain the mask image $\boldsymbol{m}_{cd}$
      \State decomposing $\boldsymbol{m}_{cd}$ into four parts, i.e., $\boldsymbol{m}_{\text{outside}}$, $\boldsymbol{m}_{\text{outside-boundary}}$, $\boldsymbol{m}_{\text{inside-boundary}}$, and $\boldsymbol{m}_{\text{inside}}$.
      \State sampling $\mathcal{N}(0, 1024)$ to derive a random vector $\boldsymbol{x} \in \boldsymbol{\mathbb{R}}^{1024\times 1}$
     \State sorting $\boldsymbol{x}$ in the descending order to derive $\boldsymbol{\hat{x}}$
     \State decomposing $\boldsymbol{\hat{x}}$ into four parts, i.e., $\boldsymbol{\hat{x}} = \{ \boldsymbol{\hat{x}}_{\text{outside}}, \boldsymbol{\hat{x}}_{\text{outside-boundary}}, \boldsymbol{\hat{x}}_{\text{inside-boundary}}, \boldsymbol{\hat{x}}_{\text{inside}}\}$
     \State placing each pixel of $\boldsymbol{\hat{x}} = \{ \boldsymbol{\hat{x}}_{\text{outside}}, \boldsymbol{\hat{x}}_{\text{outside-boundary}}, \boldsymbol{\hat{x}}_{\text{inside-boundary}}, \boldsymbol{\hat{x}}_{\text{inside}}\}$ into the above masks.
    \Until{(20,000 synthetic images are generated)}     
    \endgroup
    \Ensure  
      The synthetic dataset    
  \end{algorithmic}  
\end{algorithm}

\begin{figure}[htp]
\centering
\begin{minipage}[b]{0.99\linewidth}
\centerline{\includegraphics[scale=0.68]{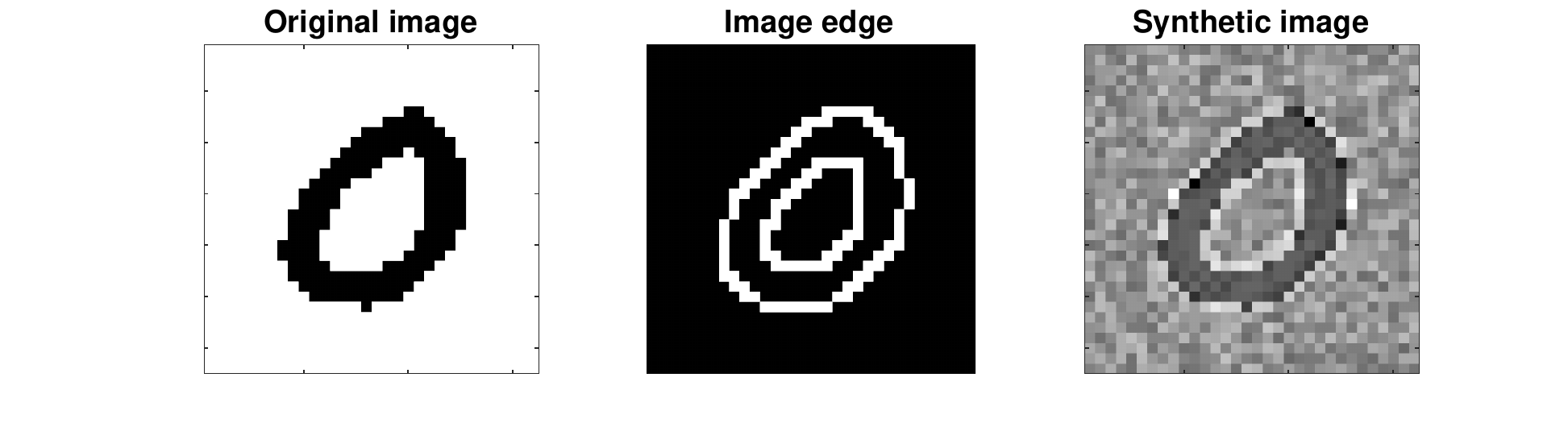}}
\end{minipage}
\vfill
\begin{minipage}[b]{0.99\linewidth}
\centerline{\includegraphics[scale=0.68]{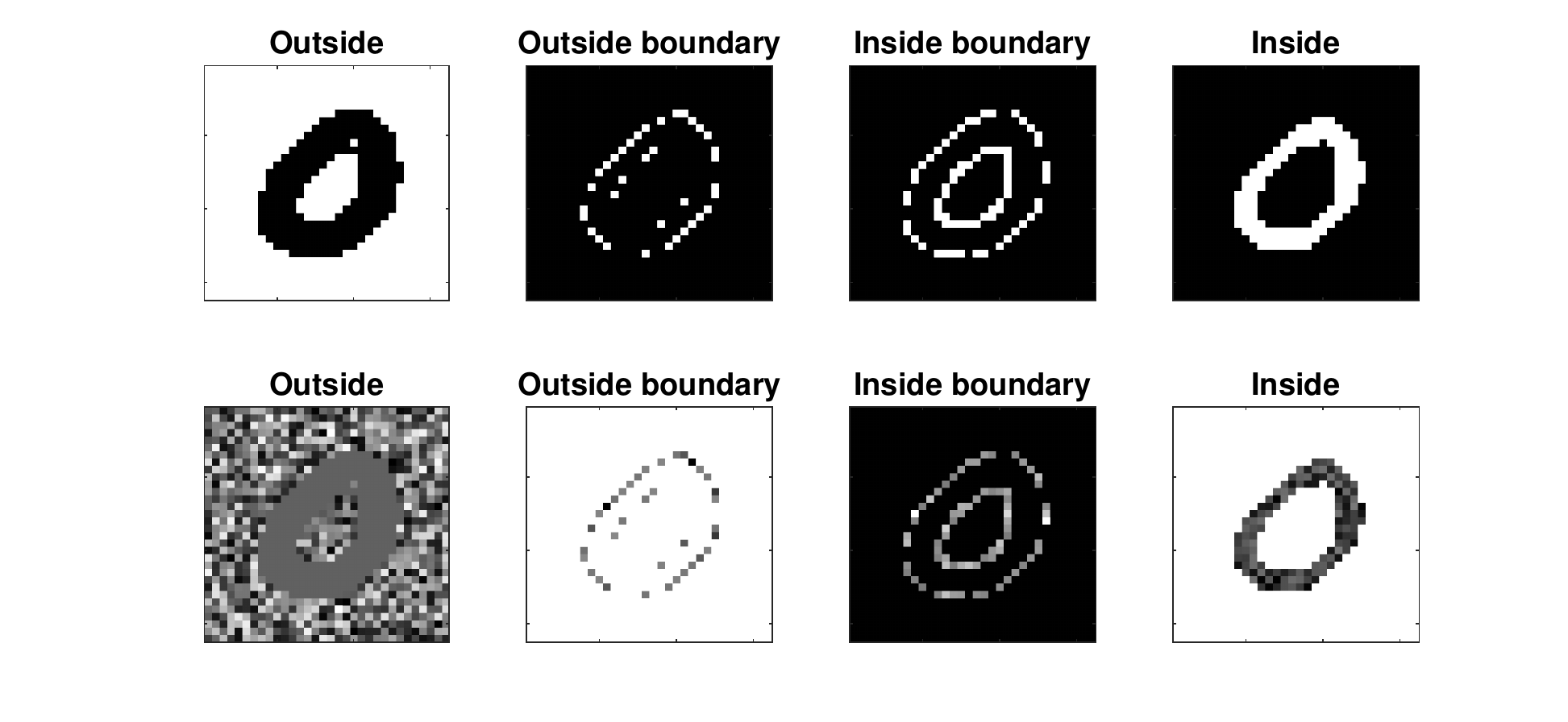}}
\end{minipage}
\caption{\small{ 
The first row shows an original image, its edge, and the corresponding synthetic image based on the original one. 
The second row uses white pixels to show the spatial position of the four mask parts, i.e., $\boldsymbol{m}_{\text{outside}}$, $\boldsymbol{m}_{\text{outside-boundary}}$, $\boldsymbol{m}_{\text{inside-boundary}}$, and $\boldsymbol{m}_{\text{inside}}$.
The third row shows the synthetic image corresponding to each mask part, i.e., $\boldsymbol{\hat{x}} = \{ \boldsymbol{\hat{x}}_{\text{outside}}, \boldsymbol{\hat{x}}_{\text{outside-boundary}}, \boldsymbol{\hat{x}}_{\text{inside-boundary}}, \boldsymbol{\hat{x}}_{\text{inside}}\}$.
}}	
\label{fig_synthetic_explanation}
\end{figure}

In this section, we present a novel approach to generate a synthetic dataset obeying the Gaussian distribution based on the NIST \footnote{\url{https://www.nist.gov/srd/nist-special-database-19}} dataset of handwritten digits.
The synthetic dataset has similar characteristics as the benchmark MNIST dataset.
It consists of 20,000 $32 \times 32$ grayscale images in 10 classes (digits from 0 to 9), and each class has 1,000 training and 1,000 testing images.
The approach to generate the synthetic dataset is summarized in Algorithm \ref{bcnn_infer}, and Figure \ref{fig_synthetic_explanation} visualize the spatial relation of $\boldsymbol{\hat{x}} = \{ \boldsymbol{\hat{x}}_{\text{outside}}, \boldsymbol{\hat{x}}_{\text{outside-boundary}}, \boldsymbol{\hat{x}}_{\text{inside-boundary}}, \boldsymbol{\hat{x}}_{\text{inside}}\}$.

\subsection{Proof: the equivalence between the gradient descent and the first order approximation}
\label{gd_first_order}

In the context of deep learning, most learning algorithms belong to the gradient descent algorithm.
Given a $\text{CNN} = \{\boldsymbol{x; f_1; ...; f_I; f_Y}\}$ and a cost function $H[\boldsymbol{f_Y}, P(\boldsymbol{Y|X})]$, where $P(\boldsymbol{Y|X})$ is the true distribution of the training label given the corresponding training dataset.
Let $\boldsymbol{\theta}$ be the parameters of the CNN,
the gradient descent aims to optimize $\boldsymbol{\theta}$ by minimizing $H[\boldsymbol{f_Y}, P(\boldsymbol{Y|X})]$ \cite{backpropagation}.
We typically update $\boldsymbol{\theta}$ iteratively to derive $\boldsymbol{\theta}^*$, which can be expressed as
\begin{equation} 
\label{gd_theta}
\boldsymbol{\theta}_{t+1} = \boldsymbol{\theta}_t - \alpha \nabla_{\theta_t} H[\boldsymbol{f_Y}, P(\boldsymbol{Y|X})]
\end{equation}
where $\nabla_{\theta_t} H[\boldsymbol{f_Y}, P(\boldsymbol{Y|X})]$ is the Jacobian matrix of $H[\boldsymbol{f_Y}, P(\boldsymbol{Y|X})]$ with respect to $\boldsymbol{\theta}_t$ at the $t$th iteration, and $\alpha > 0$ denotes a constant indicating the learning rate.
Since $P(\boldsymbol{Y|X})$ is constant, we denote $H(\boldsymbol{f_Y})$ as $H[\boldsymbol{f_Y}, P(\boldsymbol{Y|X})]$ for simplifying the following derivation. 

Since the functions of all hidden layers are differentiable and the output of a hidden layer is the input of its next layer, the Jacobian matrix of $H(\boldsymbol{f_Y})$ with respect to the parameters of the $i$th hidden layer, i.e., $\nabla_{\theta(i)} H(\boldsymbol{f_Y})$, can be expressed as follows based on the chain rule.
\begin{equation} 
\label{jacobian_theta}
\nabla_{\boldsymbol{\theta}(i)} H(\boldsymbol{f_Y}) = \nabla_{\boldsymbol{f_Y}} H(\boldsymbol{f_Y}) \cdot \nabla_{\boldsymbol{f_I}} \boldsymbol{f_Y} \cdot \prod_{j=i+1}^{I} \nabla_{\boldsymbol{f_{j-1}}} \boldsymbol{f_{j}} \cdot \nabla_{\boldsymbol{\theta}(i)}\boldsymbol{f_i}
\end{equation}
where $\boldsymbol{\theta}(i)$ denote the parameters of the $i$th hidden layer.

\begin{table}[t]
\caption{One iteration of the backpropagation training procedure for the CNN}
\label{backpropagation_dnn}
\vskip 0.15in
\begin{center}
\begin{small}
\begin{threeparttable}
\begin{tabular}{cccccc}
\toprule
Layer & Gradients update $\nabla_{\boldsymbol{\theta}_t(i)} H(\boldsymbol{f_Y})$ & & Parameters and activations update & \\
\midrule
$\boldsymbol{f_Y}$		&  ${\scriptstyle \nabla_{\boldsymbol{f_Y}} H(\boldsymbol{f_Y}) \nabla_{\boldsymbol{\theta}_t(Y)} \boldsymbol{f_Y}}$ & $\downarrow$& ${\scriptstyle \boldsymbol{\theta}_{t+1}(Y) = \boldsymbol{\theta}_{t+1}(Y) - \alpha[\nabla_{\boldsymbol{\theta}_t(Y)} H(\boldsymbol{f_Y})]}$, ${\scriptstyle \boldsymbol{f_Y}(\boldsymbol{f_2}, \boldsymbol{\theta}_{t+1}(Y))}$ & $\uparrow$ \\
$\boldsymbol{f_3}$		& ${\scriptstyle \nabla_{\boldsymbol{f_Y}} H(\boldsymbol{f_Y}) \nabla_{\boldsymbol{f_3}} \boldsymbol{f_Y} \nabla_{\boldsymbol{\theta}_t(3)} \boldsymbol{f_3}} $ & $\downarrow$& ${\scriptstyle \boldsymbol{\theta}_{t+1}(3) = \boldsymbol{\theta}_{t+1}(3) - \alpha[\nabla_{\boldsymbol{\theta}_t(3)} H(\boldsymbol{f_Y})]}$, ${\scriptstyle \boldsymbol{f_3}(\boldsymbol{f_2}, \boldsymbol{\theta}_{t+1}(3))}$ & $\uparrow$ \\
$\boldsymbol{f_2}$		& ${\scriptstyle \nabla_{\boldsymbol{f_Y}} H(\boldsymbol{f_Y}) \nabla_{\boldsymbol{f_3}} \boldsymbol{f_Y} \nabla_{\boldsymbol{f_2}} \boldsymbol{f_3} \nabla_{\boldsymbol{\theta}_t(2)} \boldsymbol{f_2}} $ & $\downarrow$& ${\scriptstyle \boldsymbol{\theta}_{t+1}(2) = \boldsymbol{\theta}_{t+1}(2) - \alpha[\nabla_{\boldsymbol{\theta}_t(2)} H(\boldsymbol{f_Y})]}$, ${\scriptstyle \boldsymbol{f_2}(\boldsymbol{f_1}, \boldsymbol{\theta}_{t+1}(2))}$ & $\uparrow$ \\
$\boldsymbol{f_1}$    	& ${\scriptstyle \nabla_{\boldsymbol{f_Y}} H(\boldsymbol{f_Y}) \nabla_{\boldsymbol{f_3}} \boldsymbol{f_Y} \nabla_{\boldsymbol{f_2}} \boldsymbol{f_3} \nabla_{\boldsymbol{f_1}} \boldsymbol{f_2} \nabla_{\boldsymbol{\theta}_t(1)} \boldsymbol{f_1}} $ & $\downarrow$& ${\scriptstyle \boldsymbol{\theta}_{t+1}(1) = \boldsymbol{\theta}_{t+1}(1) - \alpha[\nabla_{\boldsymbol{\theta}_t(1)} H(\boldsymbol{f_Y})]}$, ${\scriptstyle \boldsymbol{f_1}(\boldsymbol{x}, \boldsymbol{\theta}_{t+1}(1))}$ & $\uparrow$\\

$\boldsymbol{x}$    	        & ---	& & --- \\

\bottomrule
\end{tabular}
\begin{tablenotes}
            \item The uparrow and downarrow indicate the order of gradients and parameters(activations) update, respectively. 
\end{tablenotes}
\end{threeparttable}
\end{small}
\end{center}
\vskip -0.1in
\end{table}

Based on Equation \ref{gd_theta} and Equation \ref{jacobian_theta}, $\boldsymbol{\theta}(i)$ can be learned by the gradient descent method as
\begin{equation} 
\boldsymbol{\theta}_{t+1}(i) = \boldsymbol{\theta}_t(i) - \alpha [\nabla_{\theta_t(i)} H(\boldsymbol{f_Y})]
\end{equation}
Table \ref{backpropagation_dnn} summarizes the backpropagation training procedure for the CNN in Figure \ref{Img_cnn_prob}.

If an arbitrary function $\boldsymbol{f}$ is differentiable at point $\boldsymbol{p}^*$ in ${\mathcal{R}}^{N}$ and its differential is represented by the Jacobian matrix $\nabla_{\boldsymbol{p}}\boldsymbol{f}$, the first order approximation of $\boldsymbol{f}$ near the point $\boldsymbol{p}$ can be formulated as 
\begin{equation} 
\boldsymbol{f}(\boldsymbol{p}) - \boldsymbol{f}(\boldsymbol{p}^*) = (\nabla_{\boldsymbol{p}^*}\boldsymbol{f}) \cdot (\boldsymbol{p} - \boldsymbol{p}^*) + o(||\boldsymbol{p} - \boldsymbol{p}^*||)
\end{equation}
where $o(||\boldsymbol{p} - \boldsymbol{p}^*||)$ is a quantity that approaches zero much faster than $||\boldsymbol{p} - \boldsymbol{p}^*||$ approaches zero.

Since the gradient descent can be viewed as the first order approximation \cite{battiti1992first}, updating the activations of the hidden layers of the CNN (Figure \ref{Img_cnn_prob}) during the training procedure can be approximated as
\begin{equation} 
\label{neuron_taylor0}
\begin{split}
\boldsymbol{f_2}[\boldsymbol{f_1}, \boldsymbol{\theta}_{t+1}(2)] &\approx \boldsymbol{f_2}[\boldsymbol{f_1}, \boldsymbol{\theta}_{t}(2)] + (\nabla_{\boldsymbol{\theta}_{t}(2)} \boldsymbol{f_2}) \cdot [\boldsymbol{\theta}_{t+1}(2) - \boldsymbol{\theta}_{t}(2)] \\
\boldsymbol{f_1}[\boldsymbol{x}, \boldsymbol{\theta}_{t+1}(1)] &\approx \boldsymbol{f_1}[\boldsymbol{x}, \boldsymbol{\theta}_{t}(1)] + (\nabla_{\boldsymbol{\theta}_{t}(1)} \boldsymbol{f_1}) \cdot [\boldsymbol{\theta}_{t+1}(1) -\boldsymbol{\theta}_{t}(1)] \\
\end{split}
\end{equation}
where $\boldsymbol{f_2}[\boldsymbol{f_1}, \boldsymbol{\theta}_{t}(2)]$ denote the activations of the second hidden layer based on the parameters learned in the $t$th iteration, i.e., $\boldsymbol{\theta}_{t}(2)$, given the activations of the first hidden layer, i.e., $\boldsymbol{f_1}$.
The definitions of $\boldsymbol{f_2}[\boldsymbol{f_1}, \boldsymbol{\theta}_{t+1}(2)]$ and $\boldsymbol{f_1}(\boldsymbol{x}, \boldsymbol{\theta}_{t}(1))$ are the same as $\boldsymbol{f_2}[\boldsymbol{f_1}, \boldsymbol{\theta}_{t}(2)]$.

Since $\boldsymbol{f_2}$ has $Q$ convolutional channels, i.e.,$\boldsymbol{f_{2}} = \{\boldsymbol{f^q_{2}} = \sigma_2(\sum_{n=1}^N \boldsymbol{S^{(q, n)}_{2}} \circ \boldsymbol{f^n_1} + b^q_{2}\cdot \boldsymbol{1})\}_{q=1}^Q$, the parameters of a single convolutional channel can be expressed as $\boldsymbol{\theta^q}(2) = \{\boldsymbol{S^{(q, 1)}_{2}}, \cdots, \boldsymbol{S^{(q, N)}_{2}}, b^q_2 \}$.
As a result, $\nabla_{\boldsymbol{\theta^q}_t(2)} \boldsymbol{f^q_2}$ can be expressed as
\begin{equation} 
\begin{split}
\nabla_{\boldsymbol{\theta^q}_t(2)} \boldsymbol{f^q_2} =  (\nabla_{\sigma_2} \boldsymbol{f^q_2}) \cdot [\boldsymbol{f^1_1}; \cdots, \boldsymbol{f^N_1}, 1]^T
\end{split}
\end{equation}
where $\nabla_{\sigma_2} \boldsymbol{f^q_2} = \frac{\partial \boldsymbol{f^q_2}[\boldsymbol{f_1}, \boldsymbol{\theta}_t(2)]}{\partial \sigma_2}$. 
If we only consider a single convolutional channel $\boldsymbol{f^q_2}$, its activations updating can be approximated as
\begin{equation} 
\label{neuron_taylor0}
\begin{split}
\boldsymbol{f^q_2}[\boldsymbol{f_1}, \boldsymbol{\theta^q}_{t+1}(2)] \approx \boldsymbol{f^q_2}[\boldsymbol{f_1}, \boldsymbol{\theta^q}_{t}(2)] + (\nabla_{\boldsymbol{\theta^q}_{t}(2)} \boldsymbol{f^q_2}) \cdot [\boldsymbol{\theta^q}_{t+1}(2) - \boldsymbol{\theta^q}_{t}(2)] \\
\end{split}
\end{equation}
Substituting $(\nabla_{\sigma_2} \boldsymbol{f^q_2}) \cdot [\boldsymbol{f^1_1}; \cdots, \boldsymbol{f^N_1}, 1]^T$ for $\nabla_{\boldsymbol{\theta^q}_{t}(2)} \boldsymbol{f^q_2}$ in Equation \ref{neuron_taylor0}, we can derive 
\begin{equation}
\label{neuron_taylor}
\begin{split}
\boldsymbol{f^q_2}[\boldsymbol{f_1}, \boldsymbol{\theta}_{t+1}(2)] \approx & \boldsymbol{f^q_2}[\boldsymbol{f_1}, \boldsymbol{\theta}_t(2)] + (\nabla_{\sigma_2} \boldsymbol{f^q_2}) \cdot [\boldsymbol{f^1_1}; \cdots, \boldsymbol{f^N_1}, 1]^T \cdot \boldsymbol{\theta^q}_{t+1}(2) \\
&- (\nabla_{\sigma_2} \boldsymbol{f^q_2}) \cdot [\boldsymbol{f^1_1}; \cdots, \boldsymbol{f^N_1}, 1]^T \cdot \boldsymbol{\theta^q}_t(2)
\end{split}
\end{equation}

Assuming $\boldsymbol{\theta^q}_{t+1}(2) = \{\boldsymbol{S^{(q, 1)}_{2}}, \cdots, \boldsymbol{S^{(q, N)}_{2}}, b^q_2 \}$ and $\boldsymbol{\theta^q}_t(2) = \{ \boldsymbol{S^{(q, 1)}_{2*}}, \cdots, \boldsymbol{S^{(q, N)}_{2*}}, b^q_{2*} \}$, thus $[\boldsymbol{f^1_1}; \cdots, \boldsymbol{f^N_1}, 1]^T \cdot \boldsymbol{\theta^q}_{t+1}(2) = \sum_{n=1}^N \boldsymbol{S^{(q, n)}_{2}} \circ \boldsymbol{f^n_1} + b^q_{2}\cdot \boldsymbol{1}$ because a convolution operation is linear.
As a result, Equation \ref{neuron_taylor} can be expressed as 
\begin{equation} 
\label{neuron_taylor2}
\begin{split}
\boldsymbol{f^q_2}[\boldsymbol{f_1}, \boldsymbol{\theta}_{t+1}(2)] \approx & (\nabla_{\sigma_2} \boldsymbol{f^q_2}) \cdot [\sum_{n=1}^N \boldsymbol{S^{(q, n)}_{2}} \circ \boldsymbol{f^n_1} + b^q_{2}\cdot \boldsymbol{1}] \text{ (First order approximation)} \\
& + \boldsymbol{f^q_2}[\boldsymbol{f_1}, \boldsymbol{\theta}_t(2)] - (\nabla_{\sigma_2} \boldsymbol{f^q_2}) \cdot [\sum_{n=1}^N \boldsymbol{S^{(q, n)}_{2*}} \circ \boldsymbol{f^n_1} + b^q_{2*}\cdot \boldsymbol{1}]  \text{ (Error)} \\
\end{split}
\end{equation}

Equation \ref{neuron_taylor2} indicates that $\boldsymbol{f^q_2}[\boldsymbol{f_1}, \boldsymbol{\theta}_{t+1}(2)] $ can be expressed as its first order approximation with an error component based on the activation in the previous iteration, i.e., $\boldsymbol{f^q_2}[\boldsymbol{f_1}, \boldsymbol{\theta}_{t}(2)] $.
Since $\nabla_{\sigma_2} \boldsymbol{f^q_2} = \frac{\partial \boldsymbol{f^q_2}[\boldsymbol{f_1}, \boldsymbol{\theta}_t(2)]}{\partial \sigma_2}$ is only related to $\boldsymbol{f_1}$ and the parameters in the $t$th training iteration, i.e, $\boldsymbol{\theta}_t(2)$, it can be regarded as a constant.
Also note that the error component do not contain any parameters in the $(t+1)$th training iteration. 
In summary, $\boldsymbol{f^q_2}[\boldsymbol{f_1}, \boldsymbol{\theta}_{t+1}(2)]$ can be reformulated as
\begin{equation} 
\label{neuron_taylor3}
\begin{split}
\boldsymbol{f^q_2}[\boldsymbol{f_1}, \boldsymbol{\theta}_{t+1}(2)] & \approx \boldsymbol{C_1} \cdot [\sum_{n=1}^N \boldsymbol{S^{(q, n)}_{2}} \circ \boldsymbol{f^n_1} + b^q_{2}\cdot \boldsymbol{1}]  + \boldsymbol{C_2} \\
\end{split}
\end{equation}
where $\boldsymbol{C_1} = \nabla_{\sigma_2} \boldsymbol{f^q_2}$ and $\boldsymbol{C_2} = \boldsymbol{f^q_2}[\boldsymbol{f_1}, \boldsymbol{\theta}_t(2)] - (\nabla_{\sigma_2} \boldsymbol{f^q_2}) \cdot [\sum_{n=1}^N \boldsymbol{S^{(q, n)}_{2*}} \circ \boldsymbol{f^n_1} + b^q_{2*}\cdot \boldsymbol{1}]$.
Similarly, the activations in the first hidden layer $\boldsymbol{f_1}$ also can be formulated as the first order approximation in the context of the gradient descent learning algorithm.

\subsection{Proof: the equivalence between a convolutional channel and a matrix multiplication}
\label{necessary_conv}

\begin{figure}[!t]
\centering
\includegraphics[scale=0.23]{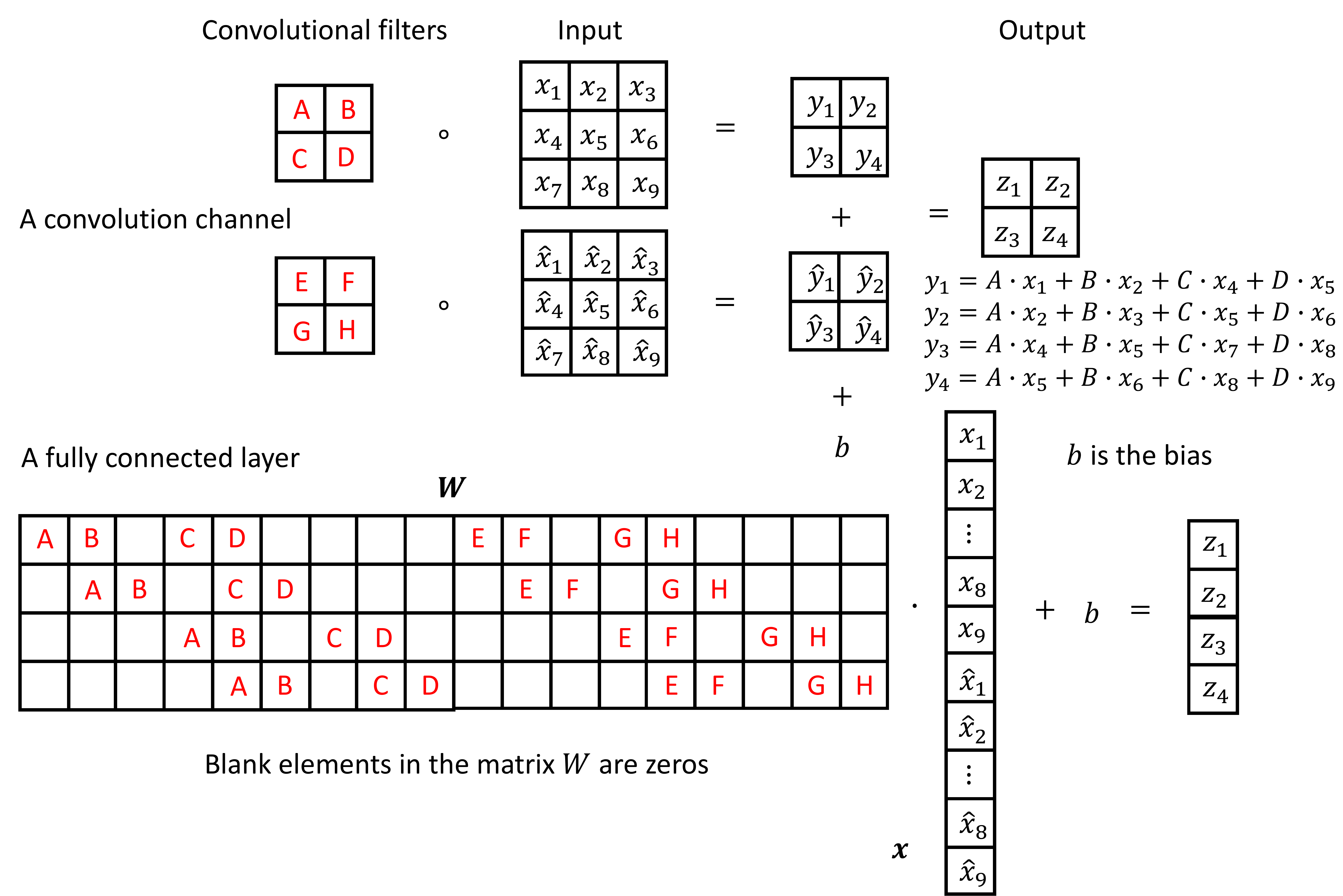}
\caption{\small{
The equivalence between a convolution channel and a matrix multiplication. 
}}
\label{Img_matrix_mulp}
\end{figure}

Assuming a $\text{CNN} = \{\boldsymbol{x; f_1; ...; f_I; f_Y}\}$ has two adjacent convolutional layers $\boldsymbol{f_i} $ and $\boldsymbol{f_{i+1}}$.
The first convolutional layer has $N$ channels, i.e. $\boldsymbol{f_i} = \{\boldsymbol{f^n_i}\}_{n=1}^N$, and the second convolutional layer has $K$ channels, i.e. $\boldsymbol{f_{i+1}} = \{\boldsymbol{f^k_{i+1}}\}_{k=1}^K$.
Therefore, the dimension of the convolutional filters $\boldsymbol{S_{i+1}}$ in $\boldsymbol{f_{i+1}}$ are $W \times H \times N \times K$, i.e., the dimension of a single filter connecting the $\boldsymbol{f^n_i}$ channel and the $\boldsymbol{f^k_{i+1}}$ channel are $D(\boldsymbol{S^{(k,n)}_{i+1}}) = W \times H$.
As a result, the $\boldsymbol{f^k_{i+1}}$ channel can be formulated as 
\begin{equation} 
\boldsymbol{f^k_{i+1}} = \sigma_{i+1}(\sum_{n=1}^N\boldsymbol{S^{(k, n)}_{i+1}} \circ \boldsymbol{f^n_i} + b^k_{i+1}\cdot \boldsymbol{1})
\end{equation}
where $\sigma_{i+1}$ is the activation function and $b_{i+1}$ is the bias for the $(i+1)$th channel.

We can regard the convolutional channel $\boldsymbol{f^k_{i+1}}$ as a matrix multiplication.
The weights of the matrix depend on the convolutional filter and the spatial location of the convolutional output, and the output of the matrix multiplication is the vectorized output of the convolutional channel.  

\pagebreak
Without considering the activation function, the equivalence between a convolutional channel and a matrix multiplication can be explained in Figure \ref{Img_matrix_mulp}, in which the first convolutional layer has two channels, i.e., $\boldsymbol{f_i} = \{\boldsymbol{x}; \boldsymbol{\hat{x}}\}$, where $\boldsymbol{x} = \{x_1, \cdots, x_9\}$ and $\boldsymbol{\hat{x}} = \{\hat{x}_1, \cdots, \hat{x}_9\}$, and the $k$th channel of the second convolutional layer is $\boldsymbol{f^k_{i+1}} = \{z_1, z_2, z_3, z_4\}$.
In addition, the dimension of the convolutional filter $\boldsymbol{S_{i+1}}$ is $2 \times 2 \times 2 \times 1$, where $\boldsymbol{S^{(k, 1)}_{i+1}} = [A,B;C,D]$ and $\boldsymbol{S^{(k, 2)}_{i+1}} = [E,F;G,H]$.
Figure \ref{Img_matrix_mulp} shows that the output of $\boldsymbol{f^k_{i+1}}$ can be expressed as a matrix multiplication. 

If we consider the activation function and make a precise statement, $\boldsymbol{f^k_{i+1}}$ can be formulated as 
\begin{equation} 
\label{conv_fc}
\boldsymbol{f^k_{i+1}} = \sigma_{i+1}(\boldsymbol{W^k_{i+1}}\cdot \boldsymbol{f_i} + b^k_{i+1} \cdot \boldsymbol{1})
\end{equation}
where $\boldsymbol{W^k_{i+1}}$ is the matrix corresponding to all the convolutional filters in the $k$th channel.




\end{document}